\documentclass[twoside]{article}

\usepackage[preprint]{aistats2026}

\usepackage{natbib}
\usepackage{graphicx}
\usepackage{algorithm,algpseudocode}
\usepackage{newfloat}
\usepackage{listings}
\usepackage[hyphens]{url}
\usepackage[breaklinks]{hyperref}
\usepackage{xcolor}
\usepackage{bm}
\usepackage{amssymb}
\usepackage{amsmath}
\usepackage{dsfont}
\usepackage{hyperref}
\usepackage{amsmath}
\usepackage{subcaption}
\usepackage{booktabs}
\usepackage{amssymb}%
\usepackage{enumitem}
\usepackage{cleveref}
\usepackage{tabularx}
\newcommand{\mat}[1]{\mathbf{#1}}

\newcommand{\sbstream}{SB-\textbf{S}ETM}
\title{Stick-Breaking Embedded Topic Model with Continuous Optimal Transport for Document Streams}
\newcommand{\softmax}{\text{softmax}}
\setlist{nolistsep}
\newcommand{\monge}{\underset{t\rightarrow t-1}{{T^\star}  }}
\begin{document}\sloppy

%
\runningtitle{{\sbstream}}

%

\twocolumn[

\aistatstitle{Stick-Breaking Embedded Topic Model with Continuous Optimal Transport for Online Analysis of Document Streams}

\aistatsauthor{ Federica Granese \And Serena Villata \And  Charles Bouveyron }

\aistatsaddress{ Université Côte d’Azur,\\CNRS, I3S, INRIA (Marianne)\\Inria Defence \& Security mission,\\France \And  Université Côte d’Azur,\\CNRS, I3S, INRIA (Marianne),\\France \And Université Côte d’Azur,\\CNRS, LJAD, INRIA (Maasai),\\France } ]

\begin{abstract}
  Online topic models are unsupervised algorithms to identify latent topics in data streams that continuously evolve over time. Although these methods naturally align with real-world scenarios, they have received considerably less attention from the community compared to their offline counterparts, due to specific additional challenges. To tackle these issues, we present {\sbstream}, an innovative model extending the Embedded Topic Model (ETM) to process data streams by merging models formed on successive partial document batches. To this end, {\sbstream} \textit{(i)} leverages a truncated stick-breaking construction for the topic–per-document distribution, enabling the model to automatically infer from the data the appropriate number of active topics at each timestep; and \textit{(ii)} introduces a merging strategy for topic embeddings based on a continuous formulation of optimal transport adapted to the high dimensionality of the latent topic space. Numerical experiments show {\sbstream} outperforming baselines on simulated scenarios. We extensively test it on a real-world corpus of news articles covering the Russian–Ukrainian war throughout 2022–2023.
\end{abstract}

\section{INTRODUCTION}
\label{sec:introduction}
Topic modeling aims to identify, in an unsupervised manner, latent topics in a collection of documents by analyzing word co-occurrence patterns. A topic is a probability distribution over the words of the vocabulary and can be interpreted through a representative set of words defining the topic. Each document can, in turn, be represented as a distribution over the topics~\citep{blei2003latent}.
Beyond their ability to analyze various forms of textual data, ranging from scientific publications to political discourse~\citep{boyd2017applications}, topic models have proven effective in several real-world applications, including customer feedback analysis, content organization, and market research for sentiment analysis and trend detection~\citep{xenos}. Many of these applications are naturally aligned with streaming settings, where new data arrives continuously over time.
This temporal dimension of topic modeling can be addressed in two main ways. The \textit{dynamic} setting assumes access to the entire historical corpus, with topic evolution analyzed retrospectively~\citep{blei2006dynamic,dieng2019dynamic,zhang2022dynamic,karakkaparambil-james-etal-2024-evaluating}. In contrast, the \textit{online} setting assumes that documents arrive sequentially, and the model must infer topics at time $t$ using only the topics inferred at time $t-1$, without revisiting past data. Despite its closer alignment with real-world scenarios, the online setting has received less attention in the literature.

A recent contribution to this setting is StreamETM~\citep{granese2025merging}, which extends the Embedded Topic Model (ETM)~\citep{dieng2020topic} by applying variational inference sequentially over consecutive batches of documents and leveraging unbalanced optimal transport to associate topics across time steps. While StreamETM performs well against strong baselines such as BERT-based models~\citep{mergebert,onlinebert}, it inherits two important limitations. First, like most topic models, it requires the number of topics $K$ to be specified a priori. This assumption is particularly problematic in streaming contexts: too few topics may prevent the discovery of new ones, while too many can lead to topic explosion. Second, its merging strategy is based on averaging topic embeddings, which may not lie in the same latent space. Although optimal transport identifies which topics should be merged, the subsequent averaging may overlook possible rotations or geometric inconsistencies in the embedding space.
In this work, we address these limitations by introducing {\sbstream}, a novel online topic modeling framework with three main contributions:

\underline{From a theoretical perspective}:
\textbf{1)} we introduce a document-level truncated stick–breaking construction in place of the ETM's logistic–normal prior, enabling data-adaptive activation of topics without fixing $K$; we provide a reparameterizable variational family via a Gaussian latent $\mathbf{z}^{(d)}$ and Kumaraswamy sticks $\bm{\nu}^{(d)}$, yielding a tractable ELBO with a dedicated KL term that controls topic activation; and \textbf{2)} we recast the task of merging models estimated at consecutive timesteps as a continuous optimal transport problem; we rely for this on an efficient low-rank computation of the Monge map for high-dimensional spaces to transport the topic embeddings of time $t$ into the latent space of time $t-1$, avoiding post-hoc averaging and preserving the geometry of the embeddings.\\
\underline{From an experimental perspective}: \textbf{3)} we evaluate {\sbstream} on an expert-annotated news corpus covering the Russian–Ukrainian war (2022–2023), showing that topic-frequency peaks align with major events. 

\begin{figure}[t]
    \centering
    \includegraphics[width=\columnwidth]{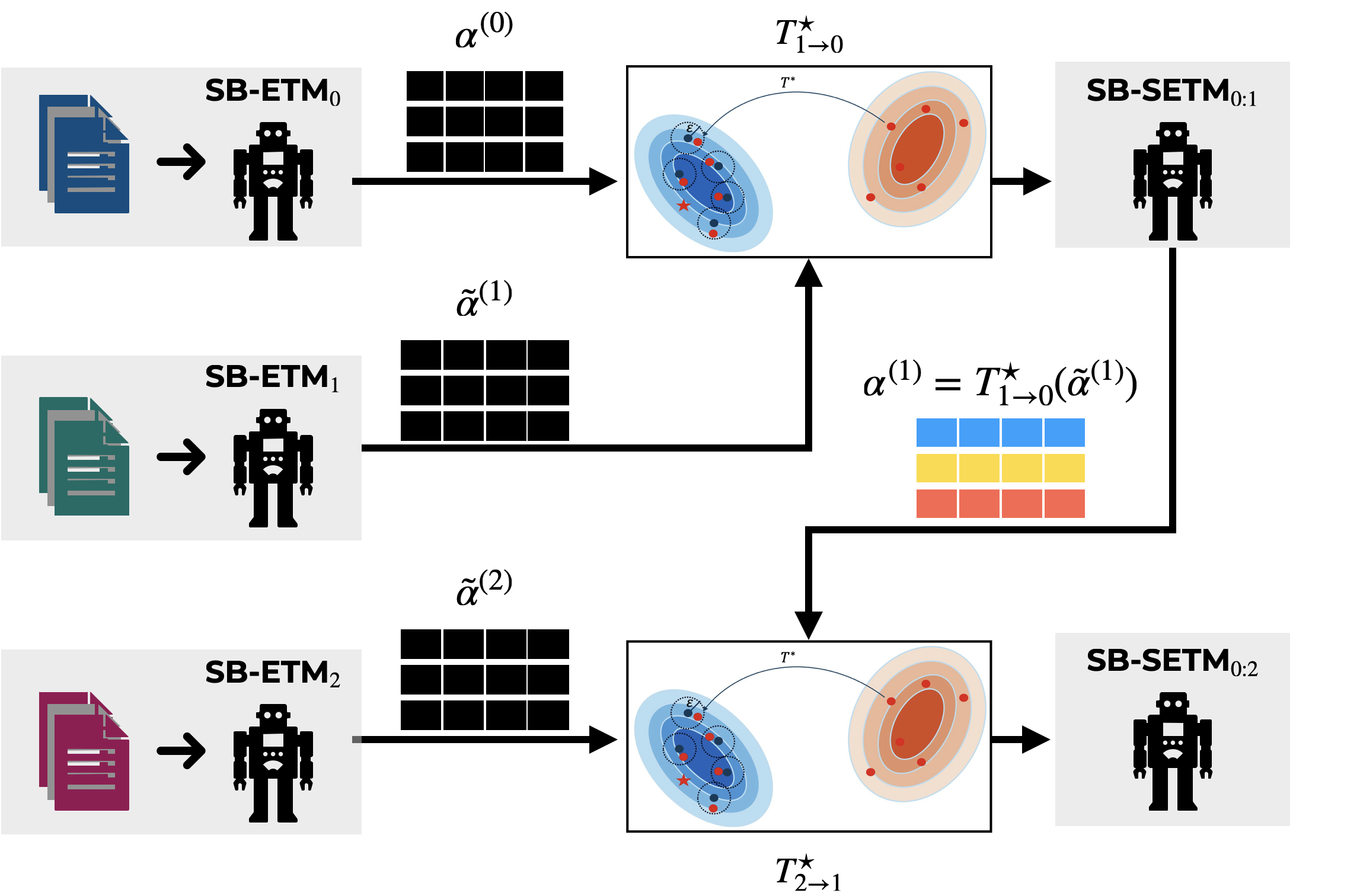}
    \caption{\textbf{The proposed {\sbstream}}. At each time step, SB-ETM$_t$ receives a new corpus of documents and infers the topic embeddings. These embeddings are then projected into the latent space of the previous timestep via the transport map $\monge$.}
    \label{fig:online_setting}
\end{figure}
\paragraph{Related works.}
OnlineLDA~\citep{hoffman2010online} is one of the first algorithms for online topic modeling based on the Latent Dirichlet Allocation model~\citep{blei2003latent}. It considers a stochastic optimization algorithm based on a natural gradient step to optimize the variational Bayes lower bound as data arrive. As OnlineLDA is not suitable for document streams that cannot be stored,~\cite{amoualian2016streaming} addresses this issue by extending LDA to document batches using copulas.
Other existing solutions for the online setting rely on a variant of BERTopic~\citep{grootendorst2022bertopic}. MergeBERT~\citep{mergebert} is a pseudo-online variant of BERTopic in which topic models are compared sequentially over time by computing the cosine similarity between topic embeddings at consecutive time steps. OnlineBERT~\citep{onlinebert} is the true online variant of the original BERTopic, preserving the embedding transformation and c-TF-IDF approach. However, it introduces online dimensionality reduction and a different clustering method, which may lead to the loss of non-linear relationships and less coherent topics in some cases.
Recently,~\cite{granese2025merging} proposed StreamETM, an extension of ETM to the online setting. The method consists of aligning topic embeddings sequentially over time using Unbalanced Optimal Transport. The merge between the topic embeddings consists of the average of the matched embeddings. As our work is a direct extension of StreamETM, which clearly outperforms online BERT approaches, we benchmark against it.

Beyond their qualities, all of these online topic models share the same two critical issues.: \textit{(i)} the number of topics must be fixed a priori, and \textit{(ii)} the consistency between the latent spaces of topic embeddings is not guaranteed when they are compared. Regarding the first issue, we remind that the number of topics might be a critical parameter of a topic model and there is no agreement in the literature on the
sequence of steps one must carry out to determine the best number of topics~\citep{bulatov2023determination}. Existing practice is to train multiple models with different values of $K$, evaluate a quality metric (often via cross-validation on held-out documents), and choose the best. This approach is not scalable in real-world online settings. Underestimating $K$ can mask topics, while overestimating $K$ tends to fragment them. 
Regarding the second issue, which is typical of MergeBERT and StreamETM, the topic embeddings at $t$ and $t-1$ could not lie in the same latent space; therefore, a mapping based on cosine similarity (as in MergeBERT) could lead to unmeaningful results. Although StreamETM accounts for this mismatch via an OT mapping, it ultimately averages the matched embeddings, a step that can be geometrically inconsistent. Therefore, it only mitigates the issue rather than solves it.

\section{THE PROPOSED APPROACH}
 \label{sec:sbstream}
\Cref{fig:online_setting} provides a graphical overview of the proposed framework. At each time step $t$, a \textbf{S}tick-\textbf{B}reaking \textbf{E}mbedded \textbf{T}opic \textbf{M}odel (SB-ETM) takes as input a corpus of documents and, for each discovered topic, produces a topic embedding $\tilde{\alpha}^{(t)}_i$. If $t>0$, we compute a transport map $\monge$ to position $\tilde{\alpha}^{(t)}_i$ into the latent space of the topics at time $t-1$, yielding $\alpha^{(t)}_i = \underset{t\rightarrow t-1}{{T^\star}  }\!\left(\tilde{\alpha}^{(t)}_i\right)$. To track topic evolution, the embeddings at $t-1$ serve as reference points: if $\alpha^{(t)}_i$ does not fall within the $\varepsilon$-neighborhood of any topic at $t-1$, we flag $\alpha^{(t)}_i$ as a new topic.
The merged model obtained by combining time steps up to $t$, uses the transported embeddings as its topic embeddings.

\textit{Remark.}
We denote with SB-ETM the model at the single timestep before any merging, i.e., the model producing the raw topic embeddings $\Tilde{\alpha}_i^{(t)}$. After merging, we refer to the resulting model as {\sbstream} (the additional S stands for \textbf{S}tream), which uses the transported embeddings $\boldsymbol{\alpha}^{(t)}$ in the reference space of $t-1$.

\subsection{SB-ETM: The Model at Time $t$}
\label{sec:Model_Timet_}
At each time step $t$, the model $M_t$ receives a corpus of documents $\mathcal{W} = \{\mathbb{W}^{(d)}\}_{d=1}^D$, where each document 
is a multiset of $N^{(d)}$ words.
Each document is then mapped to a normalized bag-of-words (BoW) vector $\mathbf{W}^{(d)} \in \mathbb{R}^V$, representing the empirical word frequency over a fixed vocabulary $\mathcal{V} = \{v_1, \dots, v_V\}$ defined a priori based on the application domain.
\paragraph{Generative model.}
We define a probabilistic decoder that generates documents from latent topic proportions. 
Each topic $k \in \{1,\dots,K\}$ is represented by an embedding $\alpha_k \in \mathbb{R}^L$ and each word $v \in \mathcal{V}$ by an embedding $\rho_v \in \mathbb{R}^L$. 
The word distribution of topic $k$ is defined as
$
\beta_k = \softmax(\bm{\rho}^\top \alpha_k) \in \Delta^V
$. 
A central limitation of classical topic models is the need to fix the number of topics $K$ in advance.  
We address this issue by adopting a truncated stick-breaking construction of the topic per document distribution $\bm\theta^d\in\Delta^{K}$. Specifically, we first draw auxiliary variables 
$\nu_k^{(d)} \sim \mathrm{Beta}(a,b), \quad k=1,\dots,K-1,\quad a,b>0$
and map them deterministically via the stick-breaking transform $f_{\text{SB}}:(0,1)^{K-1}\rightarrow\Delta^K$:
\begin{align*}
\theta_1^{(d)} &= \nu_1^{(d)},\\ 
\theta_k^{(d)} &= \nu_k^{(d)} \prod_{j=1}^{k-1}(1-\nu_j^{(d)})\quad k=2,\dots,K-1, \\
\theta_K^{(d)} &= \prod_{j=1}^{K-1}(1-\nu_j^{(d)}),
\end{align*}
where $\theta_k^{(d)}\equiv f_{\text{SB}}(\nu_k^{(d)})$.
Intuitively, the stick-breaking process consists of repeatedly breaking off and discarding a random fraction of a stick of unit length. $\nu_1^{(d)}$ determines the length of the first segment, which corresponds to the probability of the first topic; $\nu_2^{(2)}$ specifies what fraction of the remaining stick $1-\nu_1^{(d)}$ is allocated to the second topic.
Each $\nu_k^{(d)}$ therefore regulates how much probability mass is assigned to topic $k$, with later topics inheriting progressively smaller portions of the stick.
In practice, we set $K$ to a large value, allowing the model to automatically activate only as many topics as required by the data. 

Given the topic proportions $\bm{\theta}^{(d)} = \left[\theta_1^{(d)},\dots,\theta_K^{(d)}\right] \in \Delta^K$, for each token $n$ in the document $d$:
$\widehat{\tau}_{n}^{(d)}\sim\mathrm{Multinomial}(\bm{\theta}^{(d)})
$ and $\widehat{w}_{n}^{(d)} \sim \mathrm{Multinomial}(\beta_{\widehat{\tau}_{n}^{(d)}})$.
Let us denote the topic–word matrix $\bm{\beta} = (\beta_1,\dots,\beta_K) \in \mathbb{R}^{V \times K}$, marginalizing out $\widehat{\tau}_{n}^{(d)}$ we obtain:
$$p(\mathbb{W}^{(d)} \mid \bm{\theta}^{(d)}, \bm\alpha, \bm\rho) = \prod\limits_{v\in\mathcal{V}}\sum\limits_{k=1}^{K-1} \theta_k^{(d)} \, \softmax(\rho_v^\top \alpha_k). $$
\paragraph{Inference model.}
The parameters of our model are the embeddings $\bm\alpha$, and $\bm\rho$. Our goal is therefore to maximize the marginal likelihood of the documents, which, however, is intractable due to the integral over the topic proportions
\begin{align}
\label{eq:loss1}
\mathcal{L}(\bm\alpha, \bm\rho) & 
=\sum\limits_{d=1}^D \log p(\mathbb{W}^{(d)} \mid \bm\alpha, \bm\rho)  \nonumber\\
& \geq 
\sum\limits_{d=1}^D \mathbb{E}_{q_{\phi}(\bm\theta^{(d)} \mid \mathbb{W}^{(d)})}
\left[\log p(\mathbb{W}^{(d)} \mid \bm{\theta}^{(d)}, \bm\alpha, \bm\rho)\right] \nonumber\\
& \quad - \text{KL}\left(q_{\phi}(\bm\theta^{(d)} \mid \mathbb{W}^{(d)}) \mid\mid p(\bm{\theta^{(d)}})\right).
\end{align}
For completeness, we delegate in~\Cref{app:full_derivation_loss2} the derivation from the original loss to the ELBO. 

Since the Beta distribution is not easily reparameterizable, we follow the strategy in~\cite{nalisnick2016stick} and approximate it with the Kumaraswamy distribution~\citep{kumaraswamy1980generalized}.  
Crucially, this distribution has \textit{(i)} the same support as the Beta, and for equivalent parameter values of $(a,b)$, it resembles the Beta distribution, albeit with
higher entropy; \textit{(ii)} it admits a closed-form inverse CDF, which allows to generate differentiable samples by drawing $u~\sim\mathcal{U}(0,1)$ and $\nu=(1-(1-u)^{\frac{1}{b}})^{\frac{1}{a}})$; \textit{(iii)} its KL-divergence from the Beta can be approximated in closed-form for ELBO computation~\citep{nalisnick2016stick}. 

We also introduce the auxiliary variable $\mat{z}^{(d)}$ and we define it
    $\mathbf{z}^{(d)} \sim \pi_\phi(\mathbf{z}^{(d)}\mid\mathbb{W}^{(d)}) \equiv \mathcal{N}\!\big(\mathbf{z}^{(d)};\mu_\phi(\mathbb{W}^{(d)}), \sigma^2_\phi(\mathbb{W}^{(d)})\big)$, 
    $\bm{\nu}^{(d)} \sim \pi_\psi(\bm{\nu}^{(d)}\mid \mathbf{z}^{(d)}) \equiv \prod_{k=1}^{K-1}\text{Kumaraswamy}\!\left(\nu^{(d)}_k; a_\psi(\mathbf{z}^{(d)}), b_\psi(\mathbf{z}^{(d)})\right)$.

Finally, our training objective, which is a lower bound of~Eq.~\eqref{eq:loss1} (proof in~\Cref{app:full_derivation_loss_final}) is given by

\begin{align}
\label{eq:loss_final}
\widetilde{\mathcal{L}}(\bm\alpha, \bm\rho) & = 
\omega_{R}\cdot\underbrace{\sum\limits_{d=1}^D \mathbb{E}_{q_{\phi}(\bm\theta^{(d)} \mid \mathbb{W}^{(d)})}
\left[\log p(\mathbb{W}^{(d)} \mid \bm{\theta}^{(d)}, \bm\alpha, \bm\rho)\right]}_{\text{Reconstruction term}}\nonumber\\
& -\,\omega_{G}\cdot\underbrace{\text{KL}\left(\pi_\phi(\mat{z}^{(d)}\mid\mathbb{W}^{(d)})\mid\mid \mathcal{N}(0, I)\right)}_{\text{Gaussian regulirizer term}} \nonumber\\
& - \omega_{S}\cdot\underbrace{\mathbb{E}_{\pi_{\phi}}
\left [\text{KL}\left(\pi_\psi(\bm{\nu}^{(d)}\mid\mat{z}^{(d)})\mid\mid \text{Beta}(a, b)\right)\right]}_{\text{Stick-breaking term}}.
\end{align}
The reconstruction term in~Eq.~\eqref{eq:loss_final} plays the same role as in the classical ETM, ensuring that the model can accurately reconstruct the observed documents. The stick-breaking KL term regulates the effective number of active topics: for low values of $\omega_{S}$, the model tends to maintain redundant topics, whereas for high values, the model may collapse using only a few topics. Finally, the Gaussian KL term acts as a regularizer, preventing the variational posterior from drifting too far from the prior. The relative weights  $\omega_{R}$, $\omega_{G}$, $\omega_{S}$ depend also on the values of $(a,b)$ used as prior for the Beta distribution. An analysis is provided in~\Cref{app:preprocessing}. The learning algorithm is in~\Cref{alg:training-sbstream}.

\paragraph{Links with related models.}
\label{sec:comparison}
Let us quickly discuss the links and differences with related models. Regarding ETM~\cite{dieng2020topic}, the main deviation lies in the prior over document-level topic proportions: instead of the logistic–normal prior, we adopt a truncated stick–breaking prior, which naturally promotes data-adaptive allocation of topics. This modification (i) changes the generative process by replacing the logistic–normal mapping with the stick-breaking construction, and (ii) alters the training objective by introducing an additional KL term between the Kumaraswamy variational sticks and the Beta stick-breaking prior in the ELBO.
Unlike~\citet{nalisnick2016stick}, who introduce stick-breaking variational autoencoders in a general latent-variable modeling context, our approach specifically targets topic modeling, adapting the stick-breaking prior to document-level topic proportions.
Finally, we stress that our approach is conceptually different from hierarchical topic models, where the goal is to define a tree-structured organization of topics. In such models, a document is typically generated by sampling a path from the root to a leaf, selecting topics along this path, and generating words from the associated distributions~\citep{griffiths2003hierarchical}. In contrast, our goal is not to uncover explicit parent-child relations among topics at time $t$, but rather to avoid fixing the number of topics a priori, a constraint that can significantly affect performance in the online settings.


\subsubsection{Merging Consecutive Timesteps}
\label{sec:optimal}
We now focus on the way to align the models learned at two consecutive timesteps $t-1$ and $t$. 
Let us consider a stream of documents arriving as batches at discrete time steps, $\mathcal{W}_{[1:T]} = \{ \mathcal{W}^{(1)}, \dots, \mathcal{W}^{(t-1)}, \mathcal{W}^{(t)}, \mathcal{W}^{(t+1)} , \dots, \mathcal{W}^{(T)}\}$, where each $\mathcal{W}^{(i)}$ in $\mathcal{W}_{[1:T]}$ represents a corpus of documents as defined in Section \ref{sec:Model_Timet_}. Let us also assume that two SB-ETM models have been learned from the respective time batches, for which we aim at aligning the topic embeddings.
Following the approach proposed in \citep{granese2025merging}, we propose to rely on optimal transport to identify the correspondences between the identified topics of the two models and, when necessary, to attest to the discovery of new topics.  In contrast to this seminal work, which used Discrete Optimal Transport (DOT), we propose here to rely on its Continuous version (COT)~\citep{Peyre}. Crucially, instead of finding the point-to-point transport of the topic embeddings between the two models, the idea is to transport the distribution of the topic embeddings of time $t$ over the distribution of their counterparts at time $t-1$. As the topic embeddings are living in a high-dimensional latent space and knowing that OT does not scale well when the dimensionality increases, we propose to use a recently proposed approach~\citep{Bouveyron-OT}, allowing, in an efficient manner, the Continuous Optimal Transport between two high-dimensional Gaussian distributions.

Based on the modern formulation of~\citet{kantorovich1942translocation}, standard OT is generally based on the Wasserstein distance. Given two random variables $X_1$ and $X_2$ supported on $\mathbb{R}^p$, with finite second moments and whose marginal cumulative distribution functions are denoted by $\mu_1$ and $\mu_2$, respectively, the squared $2$-Wasserstein distance is defined as:
\vspace{-0.5ex}\begin{equation}
W_2^2(\mu_1, \mu_2) := \min_{\pi \in \Pi(\mu_1,\mu_2)}\mathbb{E}_{(X_1,X_2) \sim \pi} || X_1 - X_2 ||_2^2, \label{eq:Wass}
\end{equation}
where $\Pi(\mu_1, \mu_2)$ denotes the set of \emph{joint} distributions with marginals $\mu_1$ and $\mu_2$, respectively. The joint distribution $\pi^*$ minimizing Eq.~\eqref{eq:Wass} is known as an optimal coupling or optimal transport plan. Moreover, there exists a unique transport or \emph{Monge} map $T^*:\mathbb{R}^p \rightarrow \mathbb{R}^p$ linked to the optimal transport plan $\pi^*$ by the following relation:
\[
\mathbb{E}_{(X_1,X_2) \sim \pi^*}\left[ h(X_1,X_2) \right] = \mathbb{E}_{X_1 \sim\mu_1} \left[h(X_1, T^*(X_1))\right],
\]
holding for any continuous function $h:\mathbb{R}^d \times \mathbb{R}^d \rightarrow \mathbb{R}$. 
In the Gaussian case, both $W_2^2(\mu_1, \mu_2)$ and $T^*$ have closed forms~\citep{Peyre}. Yet, when the space dimensionality is large and, in particular, when the number of samples is small compared to the dimensionality, the computation errors of both quantities are extremely difficult. To tackle this issue, \cite{Bouveyron-OT}  proposed simplified closed-form expressions of the 2-Wasserstein distance and Monge transport map, with efficient and robust calculation procedures based on a low-dimensional decomposition of empirical covariance matrices. Other than analytical and computational advantages, this approach outperforms model-free methods in high dimensions, even in the case of non-Gaussian distributions. 

Assuming that $\Tilde{\alpha}^{(t)}$ and $\alpha^{(t-1)}$ are both distributed as high-dimensional ($L=300$ or $L=800$ depending on the model in the experiments) Gaussian distributions, as defined in \citep{Bouveyron-OT}, with respective intrinsic dimensionalities $d_t$ and $d_{t-1}$, it is possible to transport the topic embeddings of $t$ in the latent space of $t-1$ using the Monge map, $\forall \Tilde{\alpha}^{(t)}\in\mathbb{R}^{L}$:
$$\monge\left(\Tilde{\alpha}^{(t)}\right)=m_{t-1}+\Sigma_t^{-\frac{1}{2}} \left[ \Sigma_t^{\frac{1}{2}} \Sigma_{t-1} \Sigma_t^{\frac{1}{2}}\right]^{\frac{1}{2}} \Sigma_t^{-\frac{1}{2}} (\Tilde{\alpha}^{(t)}-m_{t}),$$
where $m_t$ and $\Sigma_t$ are respectively the mean and covariance matrix of the Gaussian distribution at time $t$, and for which $\Sigma_t^{\frac{1}{2}}$ and $\Sigma_t^{-\frac{1}{2}}$ have explicit closed-forms, cf. Theorem 2.9~\citep{Bouveyron-OT}. 
\begin{figure}
    \centering
    \includegraphics[width=0.7\columnwidth]{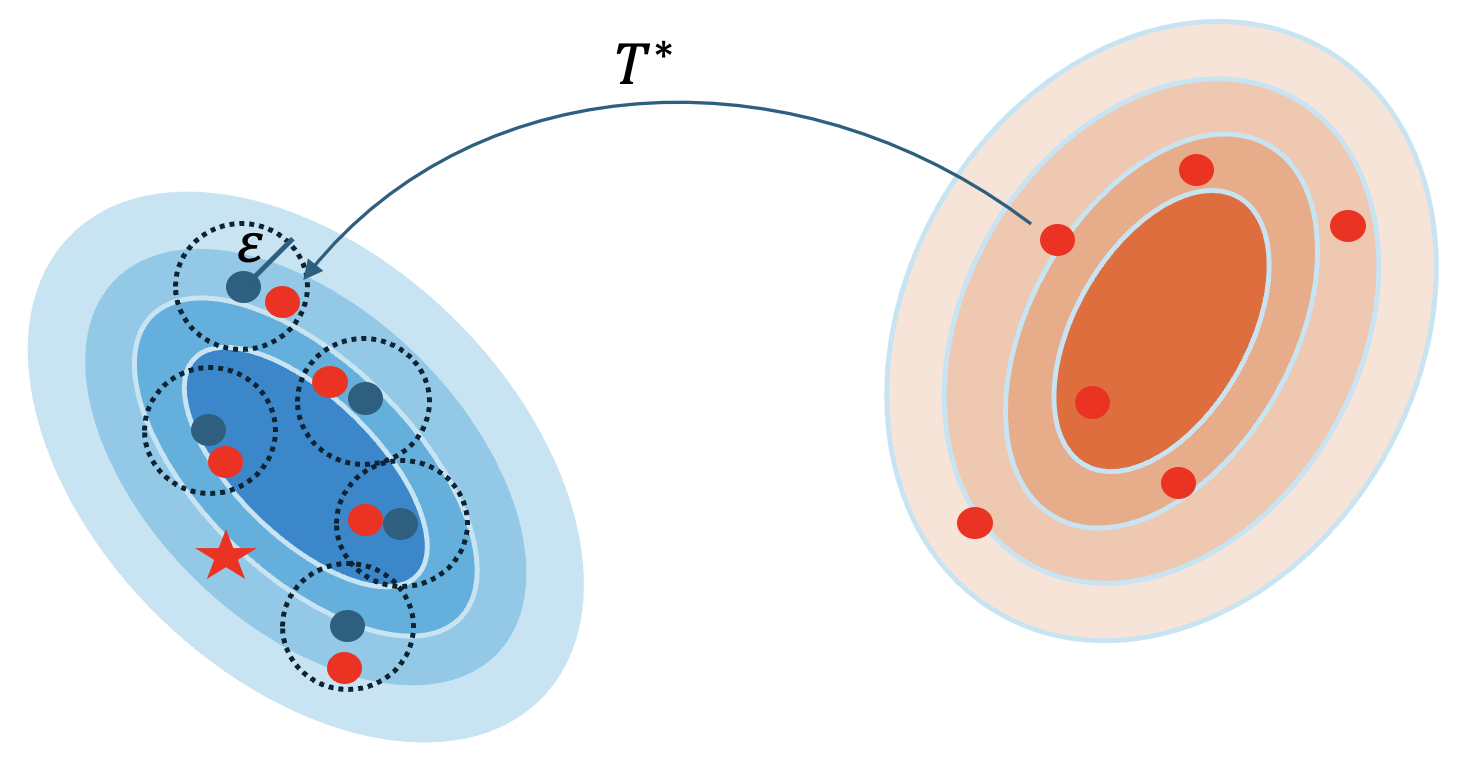}
    \caption{\textbf{Merging of topic embeddings between $t-1$ and $t$}. Latent space of $\alpha^{(t-1)}$ in blue and  of $\Tilde{\alpha}^{(t)}$ in orange. First, the topic embeddings at time $t$ are transported using the Monge map $\monge$ between two high-dimensional Gaussian distributions; then topics that are not in the $\varepsilon$-neighborhood of an existing topic will be considered as new topics (denoted by a star). }
    \label{fig:OT}
\end{figure}
\paragraph{Tracing the topic evolution.}
The COT merging approach does not allow us to directly track the evolution of topics over time, as topics are projected into embedding spaces. To address this, we propose the following strategy. Let $\alpha^{(t)} = \monge\left(\Tilde{\alpha}^{(t)}\right)$ denote the transported topic embeddings at time $t$ in the latent space of the embeddings at time $t-1$. Once the transported embeddings $\alpha^{(t)}$ are computed, we match similar topics between consecutive time steps by finding the $\varepsilon$-neighbors of each $\alpha_j^{(t-1)}$, $j = 1,
\dots,K^{(t)}$: for $\varepsilon > 0$, $\alpha_j^{(t)}$ is an $\varepsilon$-neighbor of $\alpha_j^{(t-1)}$ if $|| \alpha_j^{(t)} - \alpha_j^{(t-1)} ||_2^2 \leq \varepsilon$. After this matching, the $\alpha_j^{(t)}$ that are not associated with any $\alpha_j^{(t-1)}$ are considered new topics. \Cref{fig:OT} summarizes this merging process. Due to space limitations, additional details on the tracing strategy can be found in~\Cref{app:discussion}.


\begin{figure*}[t!]
    \centering
    \begin{subfigure}[t]{0.24\textwidth}
        \centering
        \includegraphics[width=1.05\textwidth]{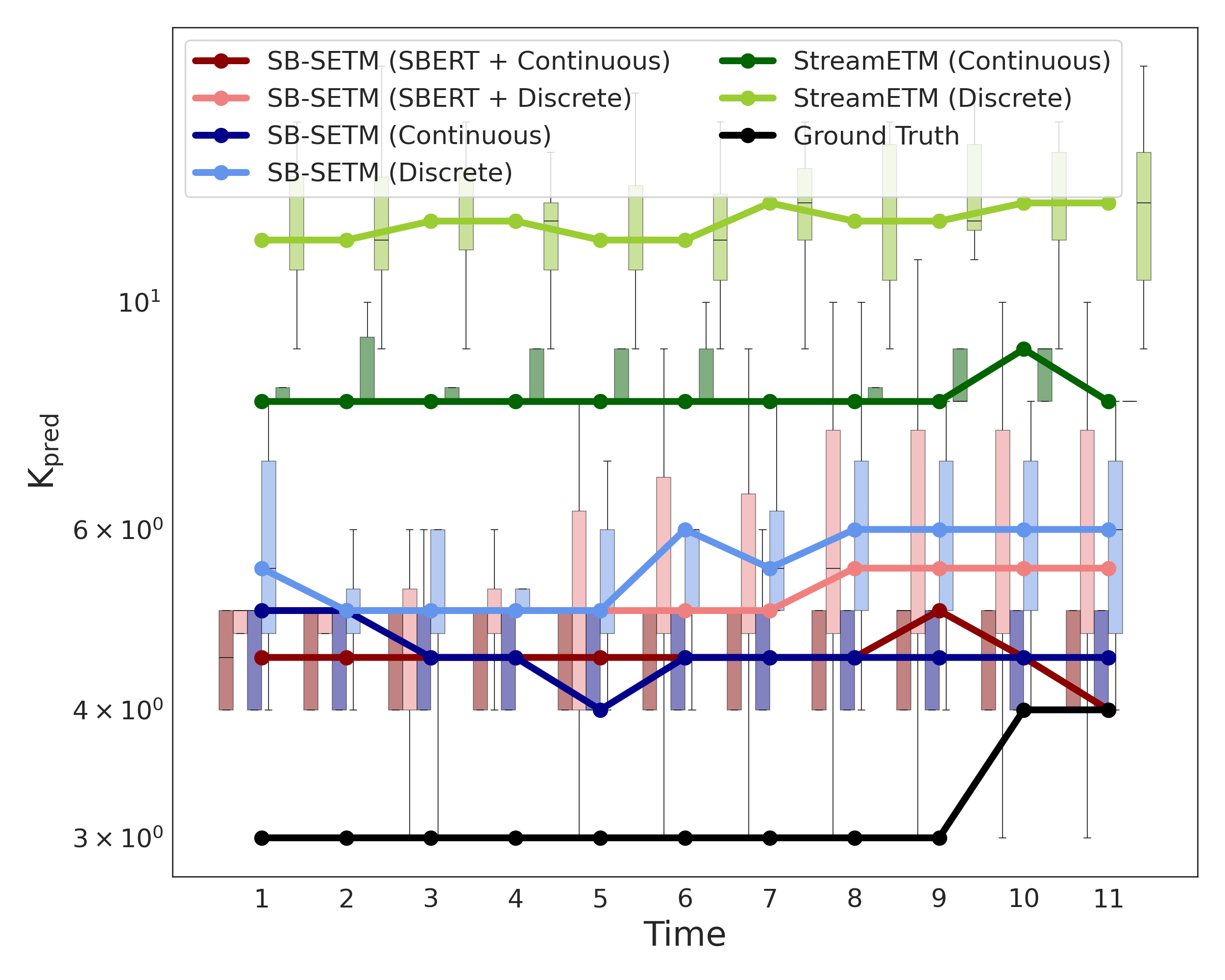}
        \caption{$K_{\texttt{init}} = 15$}
        \label{fig:15_topic}
    \end{subfigure}%
    \hfill
    \begin{subfigure}[t]{0.24\textwidth}
        \centering
        \includegraphics[width=1.05\textwidth]{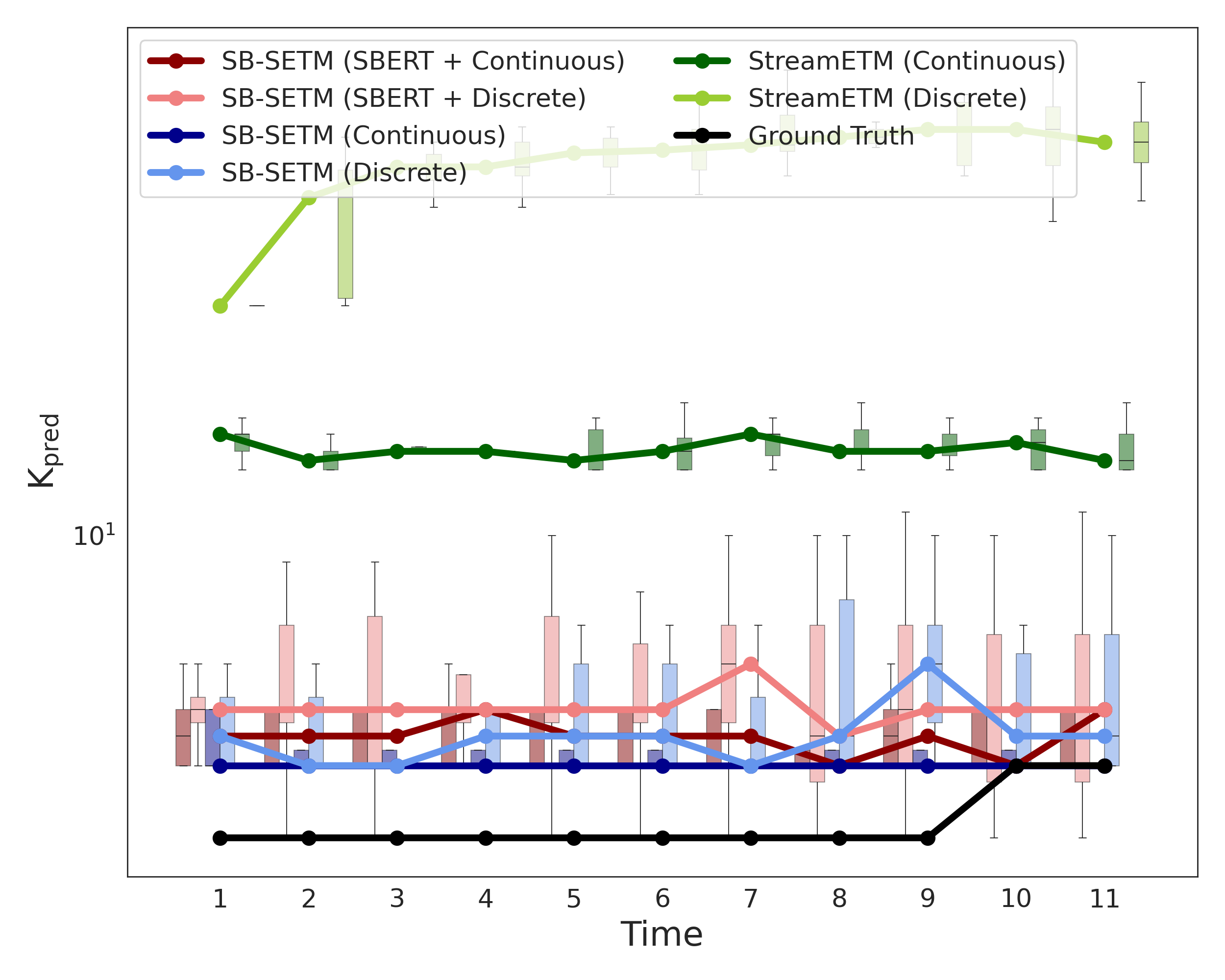}
        \caption{$K_{\texttt{init}} = 25$}
        \label{fig:25_topic}
    \end{subfigure}
    \hfill
    \begin{subfigure}[t]{0.24\textwidth}
        \centering
        \includegraphics[width=1.05\textwidth]{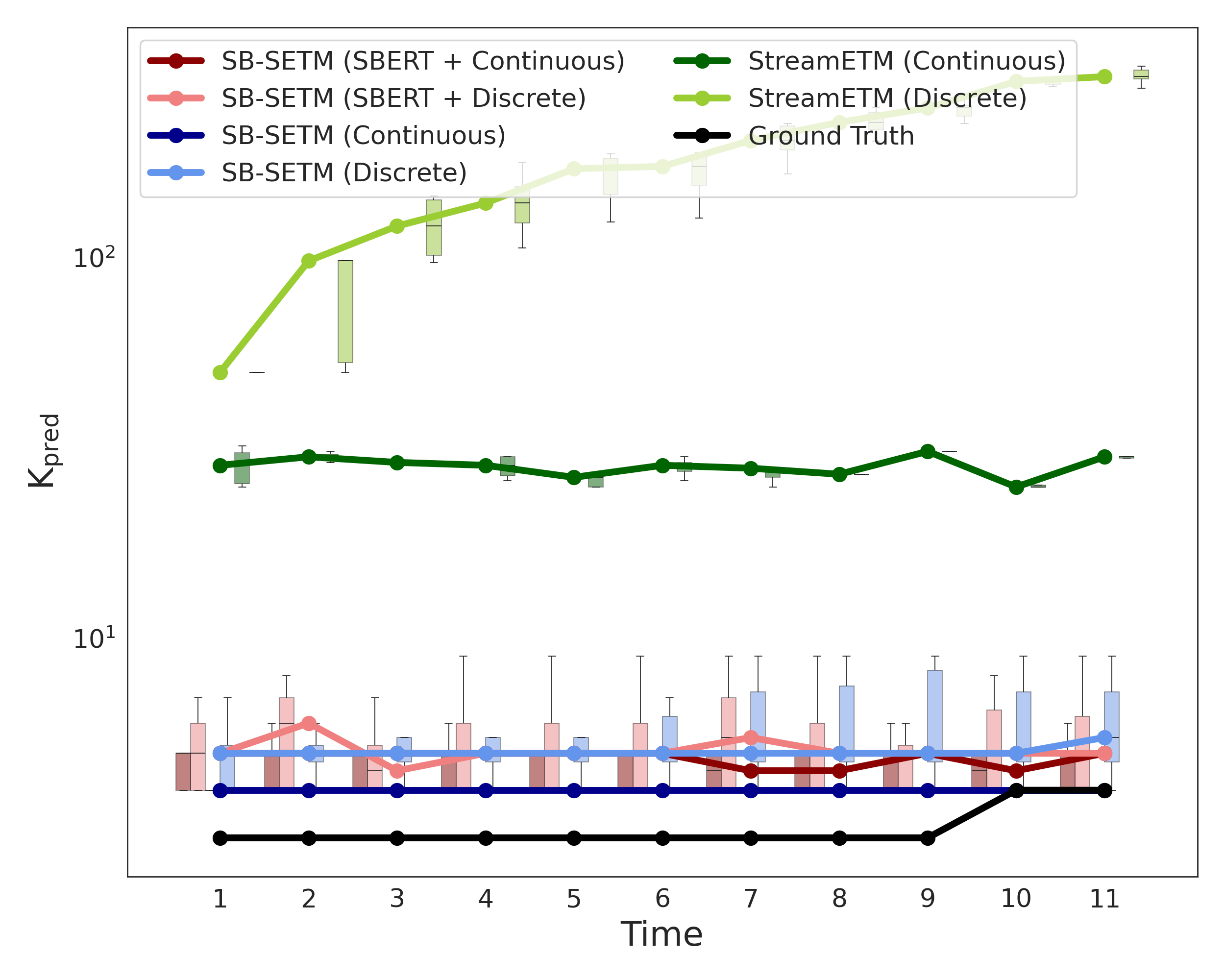}
        \caption{$K_{\texttt{init}} = 50$}
         \label{fig:50_topic}
    \end{subfigure}
    \hfill
    \begin{subfigure}[t]{0.24\textwidth}
        \centering
        \includegraphics[width=1.05\textwidth]{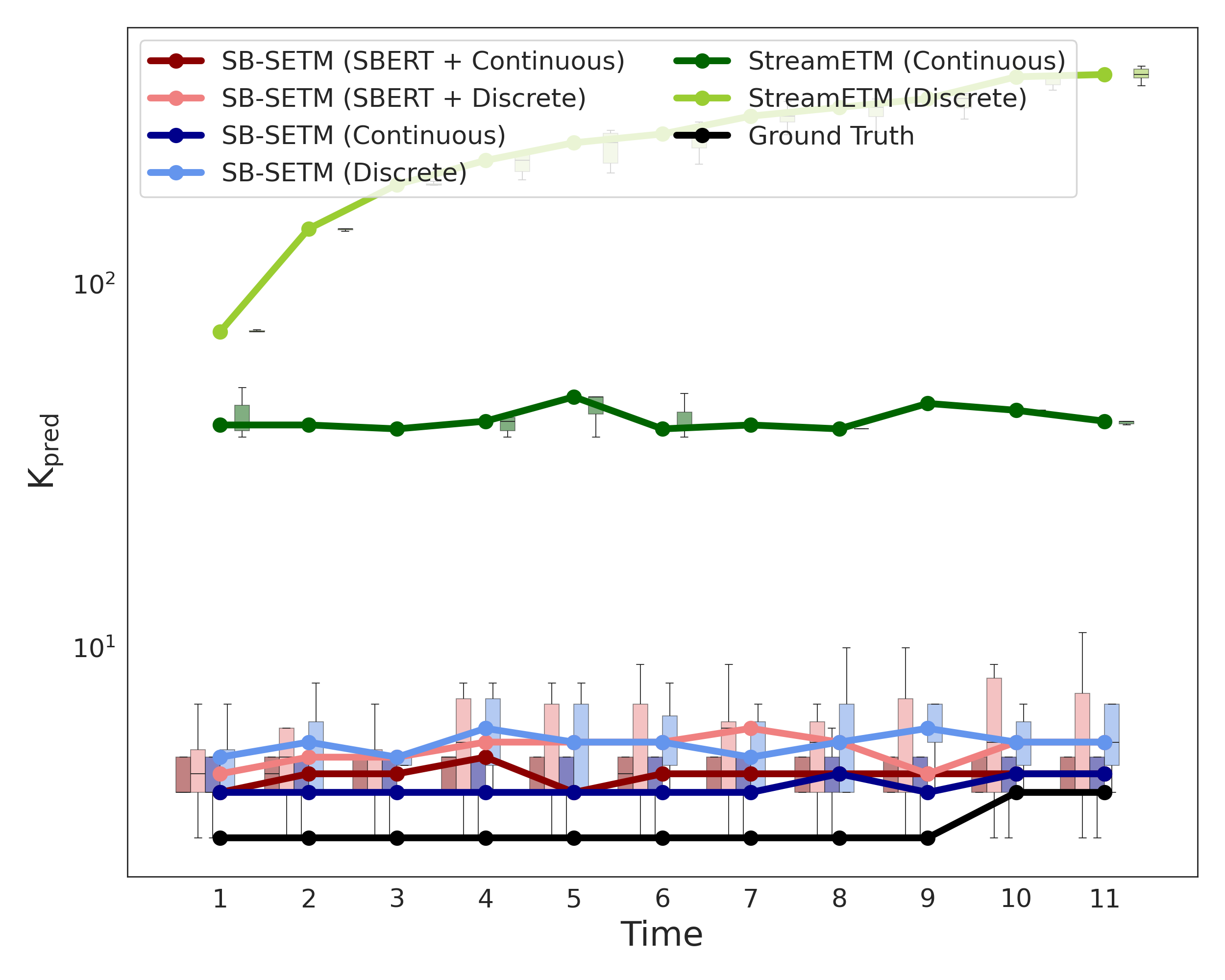}
        \caption{$K_{\texttt{init}} = 75$}
        \label{fig:75_topic}
    \end{subfigure}
    \caption{\textbf{Distribution of $K_\texttt{pred}$ (log-scale) at each timestep across 8 training runs at different $K_{\texttt{init}}$}. The solid lines connect the median values of the corresponding box plots.}
    \label{fig:n_topics}
\end{figure*}
\begin{figure*}[t!]
    \centering
    \begin{subfigure}[t]{0.25\textwidth}
        \centering
        \includegraphics[width=1.21\textwidth]{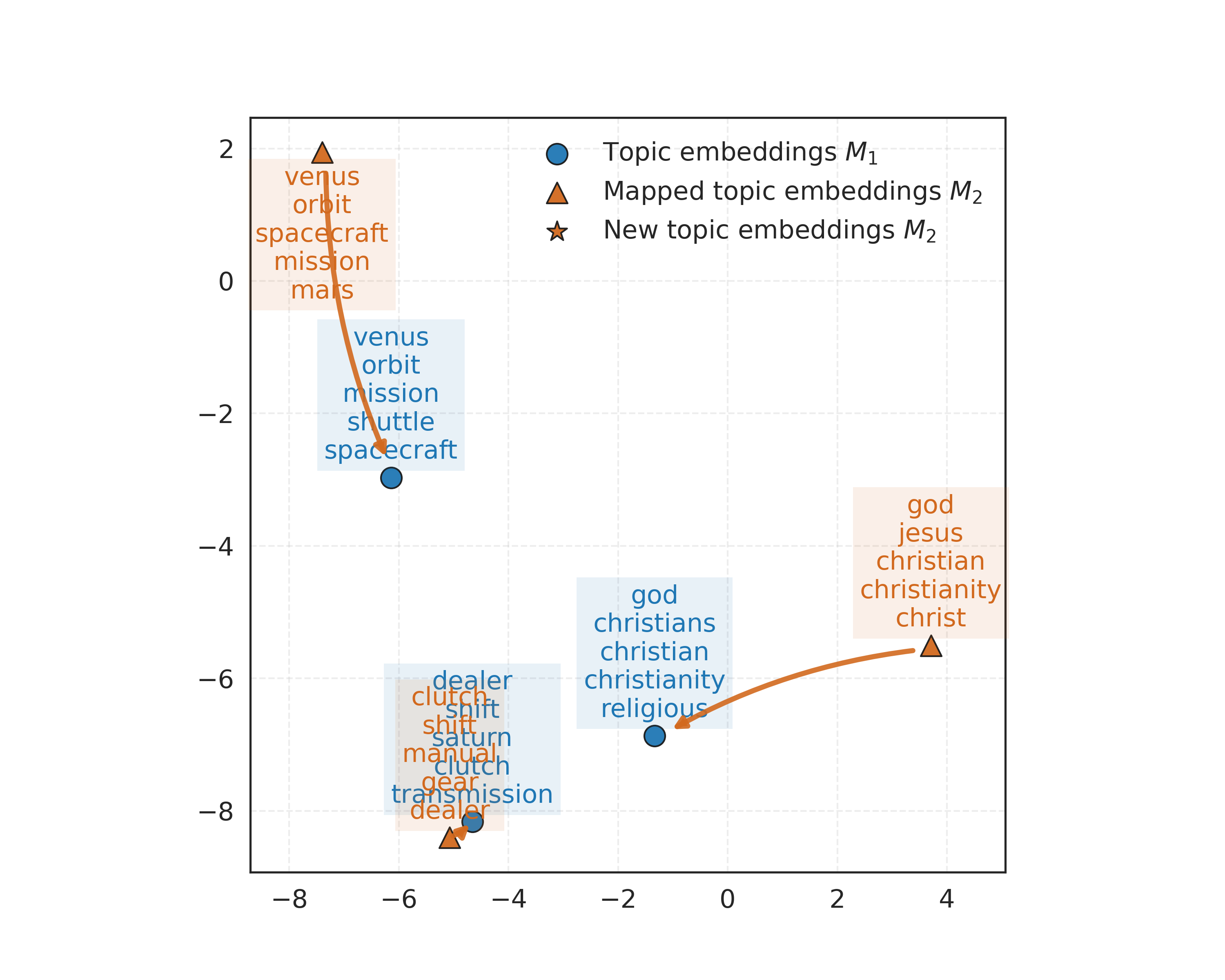}
        \caption{StreamETM}
        \label{fig:embeddings_1_original}
    \end{subfigure}%
    \hfill
    \begin{subfigure}[t]{0.25\textwidth}
        \centering
        \includegraphics[width=1.21\textwidth]{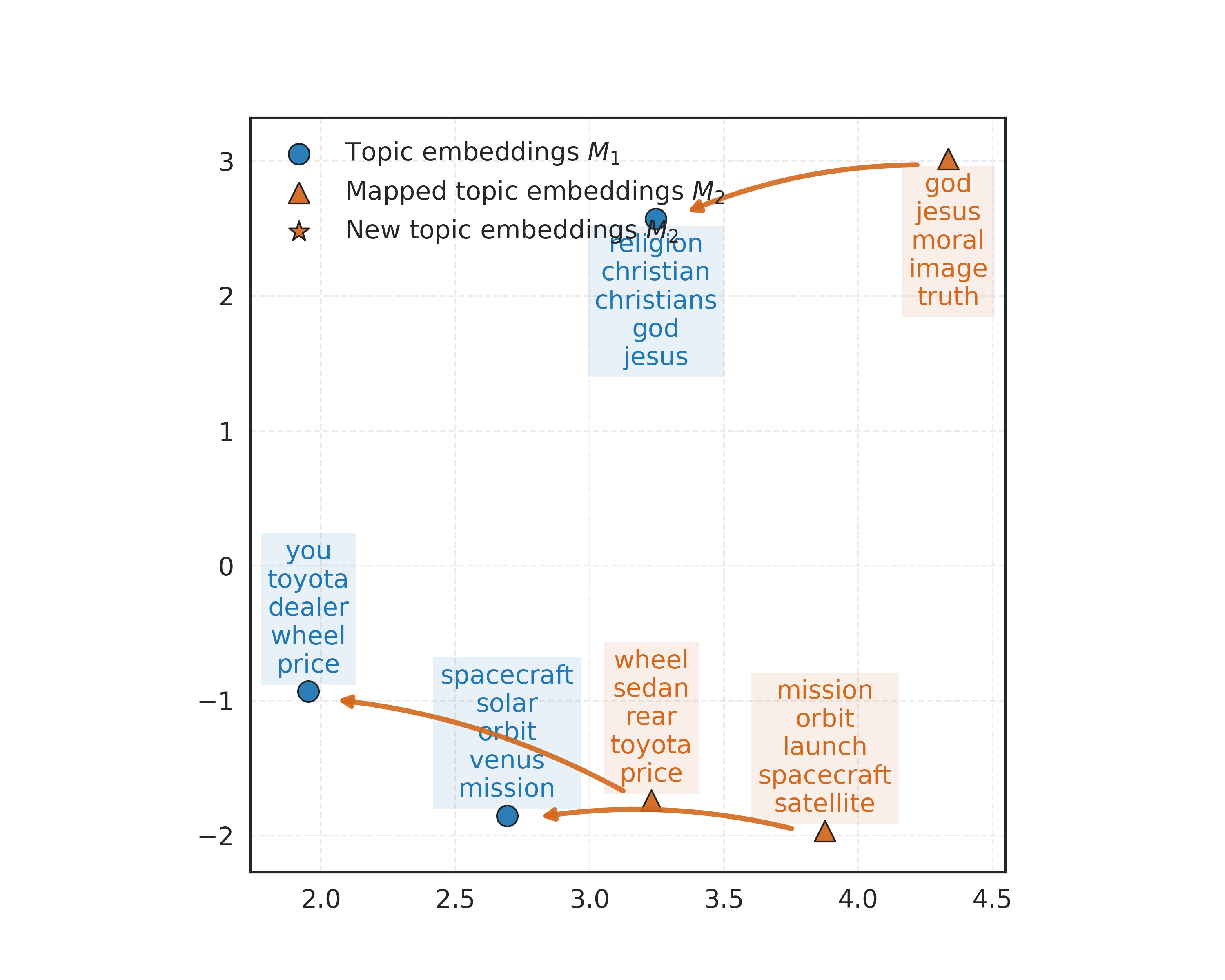}
        \caption{{\sbstream}}
        \label{fig:embeddings_1_full}
    \end{subfigure}%
    \hfill
    \begin{subfigure}[t]{0.25\textwidth}
        \centering
        \includegraphics[width=1.21\textwidth]{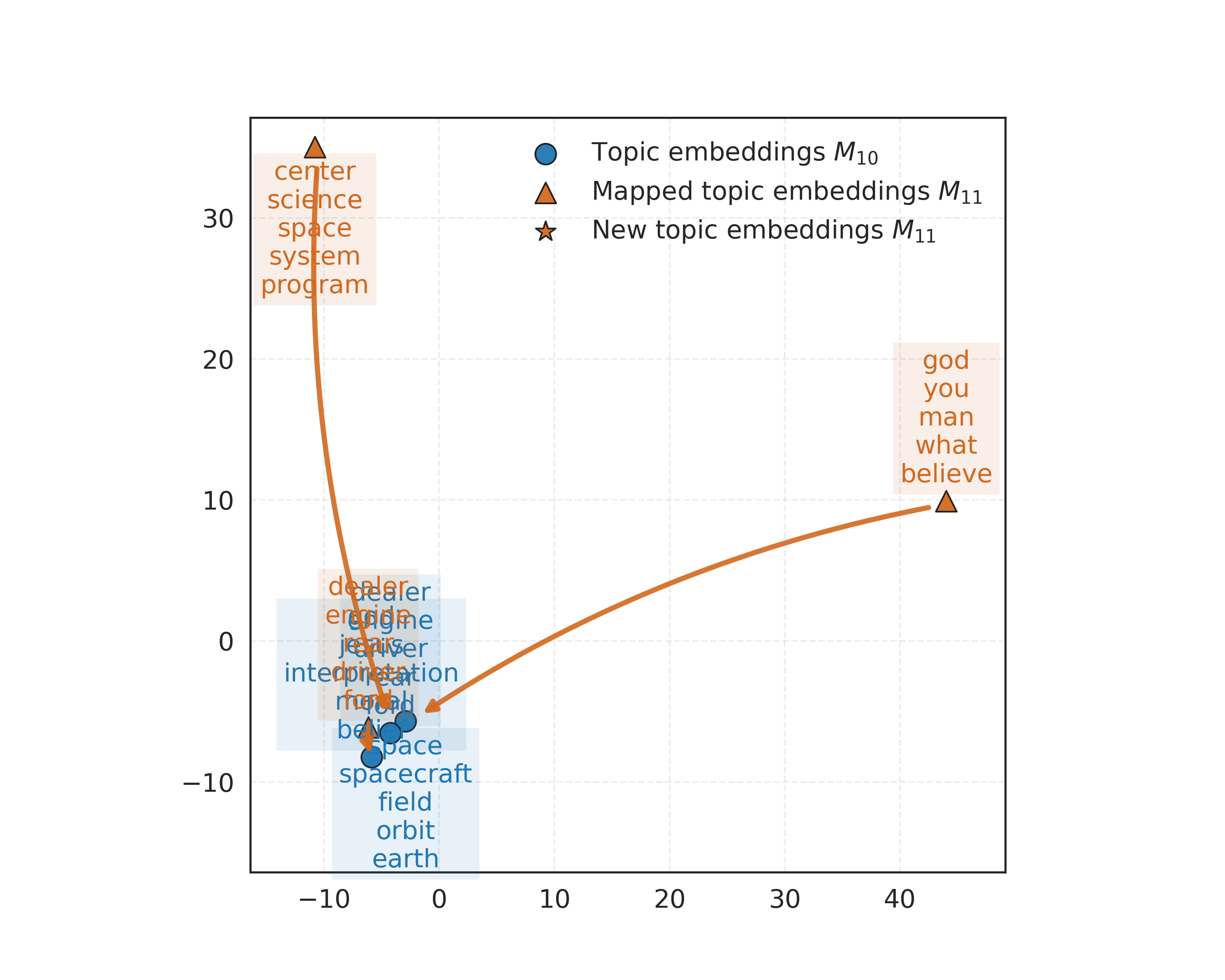}
        \caption{StreamETM}
        \label{fig:embeddings_2_original}
    \end{subfigure}%
    \hfill
    \begin{subfigure}[t]{0.25\textwidth}
        \centering
        \includegraphics[width=1.21\textwidth]{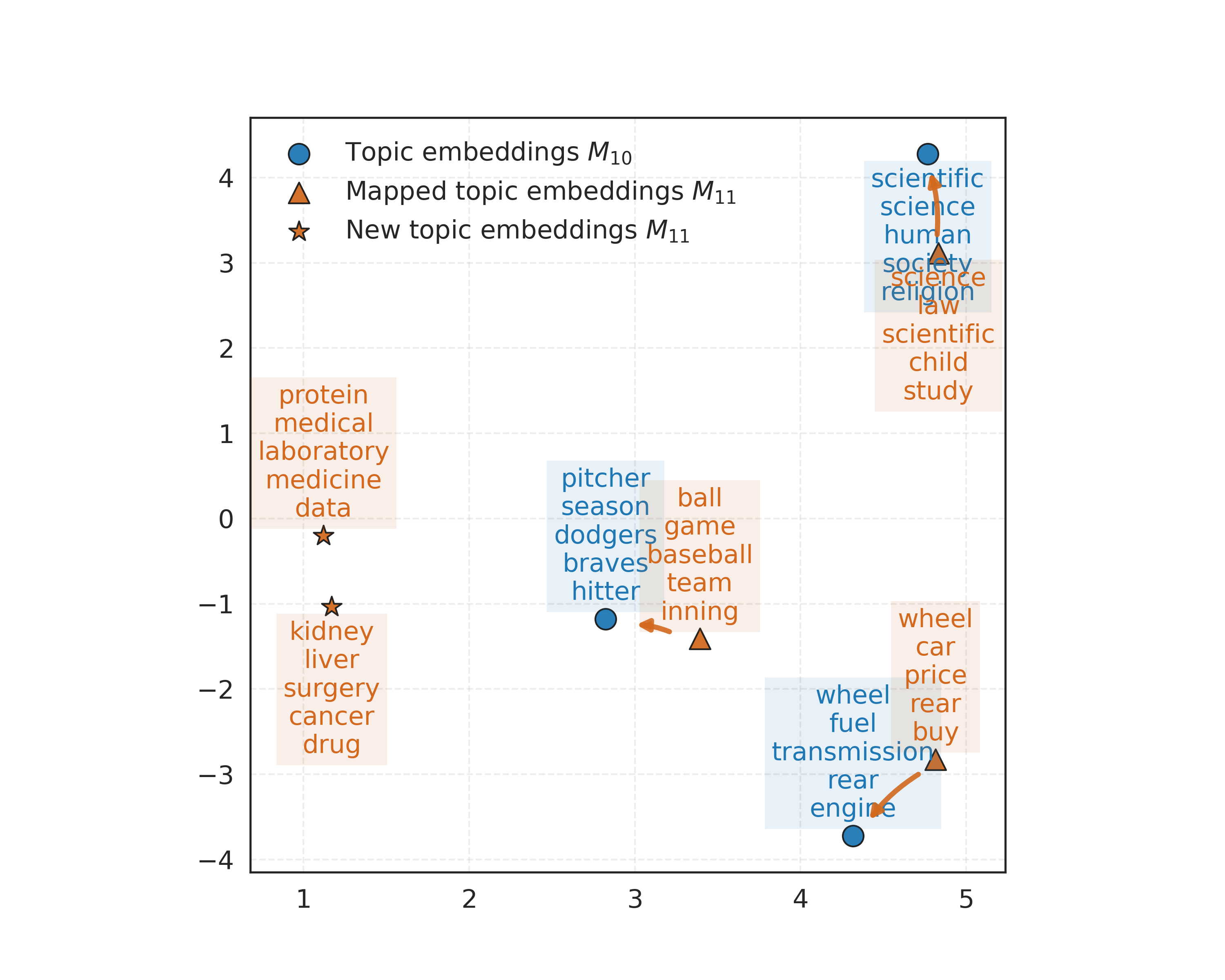}
        \caption{{\sbstream}}
         \label{fig:embeddings_2_full}
    \end{subfigure}
    \caption{\textbf{PCA projections of topic embeddings} (at timesteps $t=1$–$2$ for (a)-(b), and $t=11$–$12$ for (c)-(d)). Arrows indicate associations between topics across consecutive timesteps. In (c), the topics in blue are: god/jesus/interpretation/moral/belief, space/spacecraft/field/orbit/earth, dealer/engine/driver/rear/ford.}
    \label{fig:embeddings}
\end{figure*}
\section{EXPERIMENTAL SETTING}
\label{sec:exp_setting}
\paragraph{Datasets.} We use the \texttt{20kNewsGroupOnline} a subset of \texttt{20kNewsGroup}\footnote{\url{http://qwone.com/~jason/20Newsgroups/}} dataset designed by~\cite{granese2025merging} to mimic an online scenario. We focus on the \textsc{Custom} setting therein, in which 5 (\texttt{autos}, \texttt{sport}, \texttt{medicine}, \texttt{space}, \texttt{religion}) of the 20 topics are chosen. At each time step, at most four out of five topics are \textit{active}. Notably, at timestep 7, the topic on \texttt{space} disappears and the topic on \texttt{sport} appears. At timestep 10, the topic on \texttt{medicine} emerges.
%
We also consider the \texttt{UkRuWarNews22-23} dataset\footnote{The paper presenting the dataset is currently under submission, and cannot be referenced to ensure anonymity.}. The corpus comprises 2.652 news articles (in English and French) regarding the war Russian-Ukraine from 2022 to 2023. We focus on the 1.617 articles in English, including contributions from 113 different official news sources (e.g., president.gov.ua, Reuters, AP News, BBC, The Guardian, and CNN Edition). The articles have been categorised by three experts in political and communication sciences, over 39 distinct thematic categories on the conflict.
Datasets' details are in~\Cref{app:preprocessing}. 
\paragraph{Evaluation metrics.} We measure topic quality in terms of topic coherence (TC)~\citep{mimno2011optimizing} 
and topic diversity (TD)~\citep{dieng2020topic}. 
We report the number of predicted (active) topics ($K_{\texttt{pred}}$), and compare it with the ground-truth ($K_{\texttt{real}}$) and the initialized value ($K_{\texttt{init}}$). Ideally, $K_{\texttt{pred}}$ remains approximately constant across different $K_{\texttt{init}}$ and satisfies $K_{\texttt{pred}} \approx K_{\texttt{real}}$.
We consider $K_{\texttt{init}}\in\{15, 25, 50, 75\}$ and we let $e_i=\lvert K_{\texttt{pred}}^{(i)}-K_{\texttt{real}}\rvert$ be the topic-count
error at initialization $i$. We define the error dispersion
$\Delta=\max_{i} e_i - \min_{i} e_i$.
We combine dispersion with topic quality via the metric
$P = \Delta\cdot (1 -\text{H}_{\text{(TC, TD)}})$
where $\text{H}_{\text{(TC, TD)}}$ is the harmonic mean of TC and TD.

\paragraph{Sentence-BERT and Stick-Breaking modules.}
We modify StreamETM architecture by adding (i) document-level context via Sentence-BERT, SBERT for short, (\texttt{paraphrase-multilingual-mpnet-base-v2}) fused with BoW through self-attention, then projected to an 800-dimensional hidden space with residual blocks, batch norm, and dropout (0.1); (ii) a stick-breaking encoder with two linear layers (\texttt{fc\_a}, \texttt{fc\_b}) mapping $\mathbf{z}$ to Kumaraswamy parameters and with \texttt{softplus} to enforce $a_k,b_k > 0$.
Additional details on the model, training procedure, and computational time are provided in~\Cref{app:preprocessing}.

\begin{figure*}[t!]
        \centering
        \includegraphics[width=2\columnwidth]{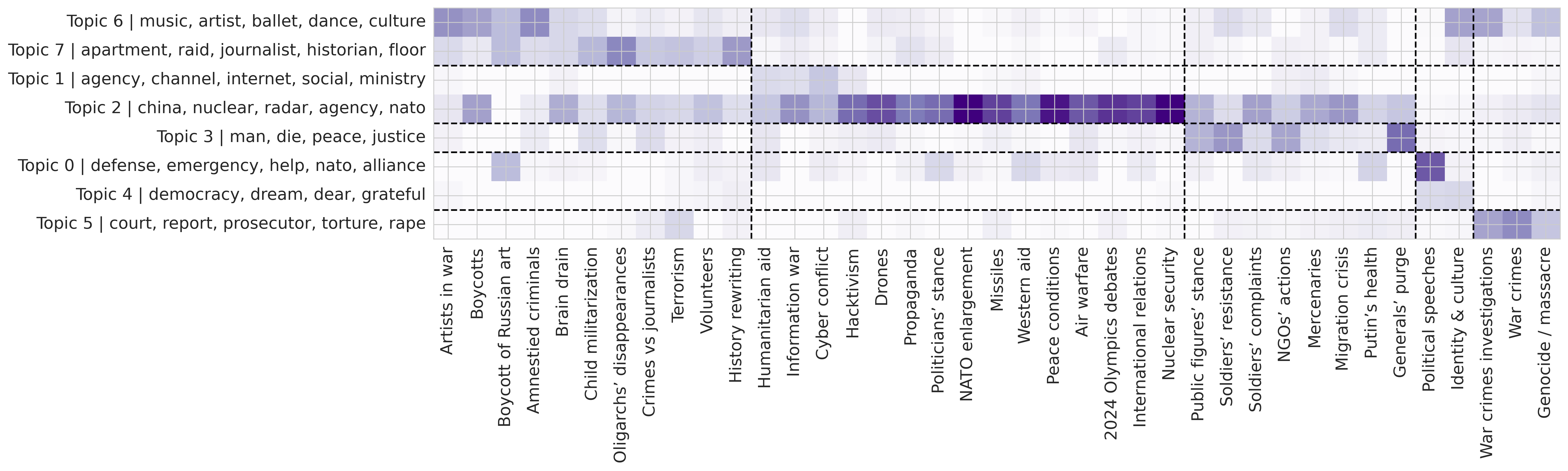}
    \caption{\textbf{Topic-term count matrix on \texttt{UkRuWarNews22-23}, {\sbstream} ($K_{\texttt{init}} = 50$).} Each row is a predicted global topic, and each column is a dataset category. The entries indicate how many documents of a given class are assigned to each topic, rescaled with inverse column weights to account for class imbalance. }
    \label{fig:cc_cigaia}
\end{figure*}

\begin{figure*}[htbp!]
        \centering        \includegraphics[width=1.9\columnwidth]{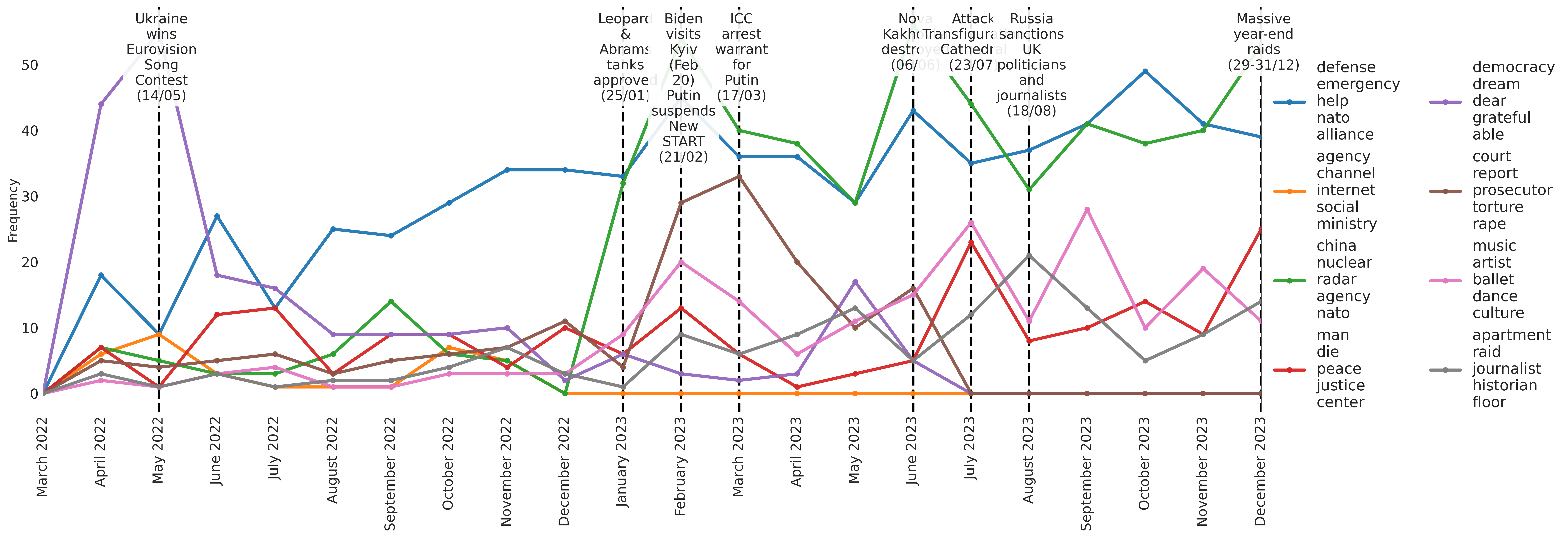}
    \caption{\textbf{Topic frequencies over time \texttt{UkRuWarNews22-23} dataset, {\sbstream} model ($K_{\texttt{init}}=50$)}.}
    \label{fig:distribution_cigaia}
\end{figure*}
\section{EXPERIMENTAL RESULTS}
\label{sec:discussion}
\begin{table}[htbp!]
    \centering
    \caption{Quantitative comparison of topic quality between the different compared approaches using the measure  $P = \Delta \cdot (1 - \text{H}_{\text{(TC, TD)}})$.}
    \begin{tabular}{r|c|c||c}
    \toprule
    Model & SBERT & OT & $P\downarrow$\\
\midrule
\textbf{{\sbstream}}  &  \textbf{$\checkmark$} & \textbf{Continuous} & \textbf{0.02}\\
{\sbstream}  &  $\checkmark$ & Discrete & 0.05\\
{\sbstream}  &  $\times$ & Continuous & 0.08\\
{\sbstream}  &  $\times$ & Discrete & 0.08\\
\midrule
\midrule
StreamETM  &  $\times$ & Continuous & 17.8\\
StreamETM  &  $\times$ & Discrete & 156.88\\
\bottomrule
\end{tabular}
\label{tab:selection}
\end{table}
Here, we distinguish between the models based on the Continuous Optimal Transport (COT-based) and Discrete Optimal Transport (DOT-based) merging strategies. Additionally, for {\sbstream}, we consider both the variant with the SBERT module and the one without it.
Full results for~\Cref{tab:quantitative} and ablations (COT vs.\ DOT; SBERT contribution) are in~\Cref{app:discussion}.
\subsection{Results for \texttt{20kNewsGroupOnline}.}
\paragraph{Choice of the number of topics.}
In~\Cref{fig:n_topics} and~\Cref{tab:quantitative}, we present the analysis of the number of predicted topics at each timestep across the 8 training iterations. This section focuses purely on a quantitative perspective, without considering the semantic content of the topics. The latter is evaluated in terms of H$_{\text{(TC, TD)}}$ in~\Cref{tab:quantitative} and further examined through the qualitative topic analysis. The black line indicates the true number of topics to be predicted. 
Ideally, a well-performing model should converge to approximately the same predicted number of topics regardless of the choice of $K_{\texttt{init}}$, indicating that it is correctly capturing the coherent semantic structure of the corpus. We observe that the {\sbstream} family of models respects this property by consistently predicting $\sim 5$ topics, regardless of the initialization. In contrast, the baseline StreamETM is highly sensitive to this parameter, predicting up to 400 topics at the last timestep when initialized with $K_{\texttt{init}}=75$. This behavior stems from the intrinsic nature of the method, which—like most topic models in the literature—requires the number of topics to be fixed in advance. Consequently, at each timestep, the model starts with that predefined number of topics, which can only increase over time. \textbf{This highlights the criticality of the choice of this parameter in an online setting}. On the other hand, when the initialization matches the true number of topics (i.e., $\sim 3$ at each timestep), StreamETM behaves more reliably. However, we emphasize that having such information in advance is highly unrealistic in real-world scenarios, as it would be equivalent to having access to the true labels in a classification task. Finally, from~\Cref{fig:n_topics}, we observe that the COT-based merging strategy yields consistently better performance across all models (shown by the darker-colored lines). Indeed, while the DOT-based merging strategy acts primarily as a mapping, where the actual merging is performed by summing the embeddings of the associated topics, in the COT case, the merge is achieved directly by transporting the topics of the current timestep into the latent space of the previous one. This reduces the number of distinct topics, as redundant topics are absorbed into a single global representation.

\paragraph{On the quality of estimated topics.}
\Cref{tab:selection} reports the numerical results for the metric $P$, which combines the harmonic mean of TC and TD with the predicted number of topics (cf.~\Cref{sec:exp_setting}). The lower, the better. From a numerical perspective, the {\sbstream} family of models behaves similarly to the variant enriched with SBERT, with the COT-based merging strategy yielding a slight improvement. The main contrast emerges with the baseline StreamETM, particularly under the DOT-based merging strategy, where performance is clearly affected by large values of $K_{\texttt{init}}$. Overall, from~\Cref{tab:quantitative} it emerges that the COT-merging strategy yields fewer topics and a higher H$_\texttt{(TC, TD)}$ no matter the
underlying model. Clearly, there is a connection between TC, TD, and $K_\texttt{pred}$. Fewer topics may reduce TD but often increase TC, and the harmonic mean balances these two effects. It is worth noting, however, that within the {\sbstream} family, the difference in the number of predicted topics between COT and DOT is relatively small (less than 2 topics on average). By contrast, the difference in terms of the harmonic mean reaches about 6 percentage points, indicating that the improvement is attributed to a genuine gain in topic quality.
\paragraph{Visualization of topic embeddings.} \Cref{fig:embeddings} shows PCA plots of topic embeddings, using a single projection computed from all embeddings for consistency.
Here, we focus on the first and second timesteps and the last two. For the baseline, we consider its best-performing configuration as in~\cite{granese2025merging} ($K_\texttt{init}=3$). For our proposed model, we select the variant yielding the best results in~\Cref{tab:selection}, i.e., the SBERT-enriched version with the COT merging strategy ($K_\texttt{init}=50$). From the plots, we highlight three main observations: \textit{(i)} topic embeddings that appear close in the Euclidean space can in fact be distant in the true latent space of the embeddings, since PCA inevitably distorts distances (e.g., in~\Cref{fig:embeddings_1_full}, wheel/sedan/rear is closer in Euclidean distance to spacecraft/solar/orbit, but COT correctly associates it with you/toyota/dealer); \textit{(ii)} StreamETM often generates topics that reuse almost the same words as in the previous timestep, a direct consequence of the DOT-merging strategy (e.g., dealer/engine/driver appears in both $t=10$ and $t=11$, and venus/orbit/spacecraft in both $t=1$ and $t=2$); \textit{(iii)}  {\sbstream} discovers the novel topics related to medicine. 

\subsection{Results for 
\texttt{UkRuWarNews22-23}}
\paragraph{On the topic-thematic association.}
Here, we focus our analysis on the best-performing model identified in the previous section, namely {\sbstream} with the COT merging strategy ($K_\texttt{init}=50$). 
The \texttt{UkRuWarNews22-23} datasets contain articles in English collected during the Russian-Ukrainian war between 2022 and 2023. In~\Cref{fig:cc_cigaia}, we visualize the reorganized topic-term count matrix, which indicates for each global topic obtained at the end, how many documents of a given category are included. The matrix has been obtained by applying the Spectral Coclustering algorithm (with 5 blocks). Because of the dataset imbalance, such counts are rescaled by applying inverse column weights, obtained as the reciprocal of the total counts per category. From a semantic perspective, the associations within the main blocks are meaningful: terms related to culture consistently cluster with categories such as Artists in the war and Boycott of Russian art; legal and judicial vocabulary (e.g., torture, rape, prosecutor) aligns with War crimes and Genocide; and technical-military lexicon (e.g., nuclear, radar, NATO) maps onto Nuclear security and International relations. More transversal categories, such as Propaganda and Information warfare, display broader activation across multiple contexts, reflecting their pervasive role in the discourse rather than methodological noise. At the same time, some themes (e.g., Hacktivism) remain marginal, which likely corresponds to their limited representation in the corpus.
\paragraph{Distribution of the topics over time.} \Cref{fig:distribution_cigaia} shows the distributions of topic frequencies over time. Since the dataset also provides a chronological list of events, we align some of the most notable events of 2023 with the observed peaks in the distributions.\\
1) \underline{January}: The United States announced the delivery of 31 Abrams tanks to Ukraine. This coincides with an increase in the blue topic, related to NATO/defense.\\
2) \underline{February}: Russia suspended its participation in New START, the last major nuclear arms control treaty with the United States, which limits deployed strategic nuclear warheads and delivery systems. We observe a rise in the green topic (nuclear issues).\\
3) \underline{March}: The International Criminal Court issued arrest warrants against Vladimir Putin and Maria Lvova-Belova for war crimes related to the deportation of Ukrainian children. This corresponds to a peak in the brown topic (reports and human rights violations).\\
5) \underline{July}: Russia adopted a law severely restricting the right to gender transition for transgender people. Later that month, Odessa was struck by drone attacks that destroyed the Cathedral of the Transfiguration. The events coincide with a rise in the red/pink topic, related to human-centered issues.\\
6) \underline{August}: Russia imposed sanctions on several British politicians and journalists in retaliation for UK support for Ukraine. This period aligns with an increase in the gray topic, connected to journalism.

The plot also reveals that some topics, while highly salient in the early stages of the conflict, gradually lost relevance over time (e.g., the democracy/dream topic). Others emerged only later, such as the green topic related to nuclear issues and China, while some remained persistent throughout the entire period, such as the blue topic connected to defense and emergency.

\section{CONCLUSION}

This work considered the challenging problem of online topic modeling for document streams. To tackle issues of previously proposed approaches, we proposed an extension of ETM, named SB-SETM. SB-SETM first enables the automatic identification of the appropriate number of active topics in each timestep, by relying on a truncated stick-breaking construction for the topic–per-document distribution. Second, SB-SETM leverages an efficient low-rank formulation of continuous optimal transport in the high-dimensional latent space of the topic embeddings to merge similar topics between two consecutive timesteps or identify new unobserved topics. SB-SETM has been shown to be more accurate both for estimating the number of topics and for providing meaningful topics on simulated and real-world scenarios. 

\bibliographystyle{apalike}
\bibliography{biblio}
\clearpage

\clearpage
\appendix
\onecolumn
\aistatstitle{Supplementary Materials}
\section{ADDITIONAL MATERIAL TO SECTION~\ref{sec:sbstream}}
\subsection{Proof of Equation~\ref{eq:loss1}}
\label{app:full_derivation_loss2}
We recall from~\Cref{sec:sbstream} the probability associated to a document $\mathbb{W}^{(d)}$ given the topic embeddings $\bm{\alpha}$ and the word embeddings $\bm{\rho}$ is
\begin{align*}
p(\mathbb{W}^{(d)} \mid \bm{\theta}^{(d)}, \bm\alpha, \bm\rho) = \prod\limits_{v\in\mathcal{V}}\sum\limits_{k=1}^{K-1} \theta_k^{(d)} \, \softmax(\rho_v^\top \alpha_k)
= \prod\limits_{v\in\mathcal{V}}\sum\limits_{k=1}^{K-1} \theta_k^{(d)} \, \beta_{v,k}.
\end{align*}
Our goal is therefore to
maximize the marginal likelihood of the documents:
\begin{align*}
\mathcal{L}(\bm\alpha,& \bm\rho) =  \sum\limits_{d=1}^D\log\int_\Theta p(\mathbb{W}^{(d)} \mid \bm{\theta}^{(d)},\bm\alpha, \bm\rho) p(\bm{\theta}^{(d)})d\bm{\theta}^{(d)}\\
= &  \sum\limits_{d=1}^D\log\int_\Theta p(\mathbb{W}^{(d)}, \bm{\theta}^{(d)} \mid \bm\alpha, \bm\rho)d\bm{\theta}^{(d)}\\
= &  \sum\limits_{d=1}^D\log\int_\Theta \frac{p(\mathbb{W}^{(d)}, \bm{\theta}^{(d)} \mid \bm\alpha, \bm\rho)q_{\phi}(\bm\theta^{(d)} \mid \mathbb{W}^{(d)})}{q_{\phi}(\bm\theta^{(d)} \mid \mathbb{W}^{(d)})}d\bm{\theta}^{(d)}
\\
= & \sum\limits_{d=1}^D\log \mathbb{E}_{q_{\phi}(\bm\theta^{(d)} \mid \mathbb{W}^{(d)})}\left[
\frac{p(\mathbb{W}^{(d)}, \bm{\theta}^{(d)} \mid \bm\alpha, \bm\rho)}{q_{\phi}(\bm\theta^{(d)} \mid \mathbb{W}^{(d)})}\right]
\\
\geq &  \sum\limits_{d=1}^D \mathbb{E}_{q_{\phi}(\bm\theta^{(d)} \mid \mathbb{W}^{(d)})}\left[\log
\frac{p(\mathbb{W}^{(d)}, \bm{\theta}^{(d)} \mid \bm\alpha, \bm\rho)}{q_{\phi}(\bm\theta^{(d)} \mid \mathbb{W}^{(d)})}\right] \quad\quad\quad\quad\text{(Jensen inequality)}
\\
= & \sum\limits_{d=1}^D \mathbb{E}_{q_{\phi}(\bm\theta^{(d)} \mid \mathbb{W}^{(d)})}
\left[\log p(\mathbb{W}^{(d)}, \bm{\theta}^{(d)} \mid \bm\alpha, \bm\rho)\right] -
\mathbb{E}_{q_{\phi}(\bm\theta^{(d)} \mid \mathbb{W}^{(d)})}\left[\log q_{\phi}(\bm\theta^{(d)} \mid \mathbb{W}^{(d)})\right] \\
= & \sum\limits_{d=1}^D \mathbb{E}_{q_{\phi}(\bm\theta^{(d)} \mid \mathbb{W}^{(d)})}
\left[\log p(\mathbb{W}^{(d)}, \bm{\theta}^{(d)} \mid \bm\alpha, \bm\rho)\right] + \mathcal{H}\left(q_{\phi}(\bm\theta^{(d)} \mid \mathbb{W}^{(d)})\right)\\
= & \sum\limits_{d=1}^D \mathbb{E}_{q_{\phi}(\bm\theta^{(d)} \mid \mathbb{W}^{(d)})}
\left[\log p(\mathbb{W}^{(d)} \mid \bm{\theta}^{(d)}, \bm\alpha, \bm\rho)\right] + 
\mathbb{E}_{q_{\phi}(\bm\theta^{(d)} \mid \mathbb{W}^{(d)})}
\left[\log p(\bm{\theta}^{(d)} )\right] +
\mathcal{H}\left(q_{\phi}(\bm\theta^{(d)} \mid \mathbb{W}^{(d)})\right)\\
= & \sum\limits_{d=1}^D \mathbb{E}_{q_{\phi}(\bm\theta^{(d)} \mid \mathbb{W}^{(d)})}
\left[\log p(\mathbb{W}^{(d)} \mid \bm{\theta}^{(d)}, \bm\alpha, \bm\rho)\right] - \text{KL}\left(q_{\phi}(\bm\theta^{(d)} \mid \mathbb{W}^{(d)}) \mid\mid p(\bm{\theta}^{(d)})\right),
\end{align*}
where $\mathcal{H}(\cdot)$ represents the entropy and $\text{KL}(\cdot)$ is the Kullback-Leibler divergence.

\subsection{Proof of Equation~\ref{eq:loss_final}}
\label{app:full_derivation_loss_final}
We first recall the stick-breaking transformation $f_{\mathrm{SB}}:(0,1)^{K-1}\rightarrow \Delta^{K-1}$, which deterministically maps a sequence of auxiliary variables $\bm{\nu}^{(d)}=(\nu^{(d)}_1,\dots,\nu^{(d)}_{K-1}) \in (0,1)^{K-1}$ into a valid topic proportion vector $\bm{\theta}^{(d)}\in\Delta^{K-1}$. Intuitively, each $\nu^{(d)}_k$ specifies the fraction of the remaining stick that is assigned to the $k$-th component, while the last component $\theta^{(d)}_K$ collects the residual mass. Hence,
\[
\bm\theta^{(d)} = f_{\text{SB}}(\bm\nu^{(d)}).
\]

The latent variable $\mathbf{z}^{(d)}$ does not directly enter the transformation, but it controls the distribution of $\bm\nu^{(d)}$ through a conditional variational distribution. Concretely, we define
\begin{align*}
    \mathbf{z}^{(d)} &\sim \pi_\phi(\mathbf{z}^{(d)}\mid\mathbb{W}^{(d)}) \equiv \mathcal{N}\!\big(\mathbf{z}^{(d)};\mu_\phi(\mathbb{W}^{(d)}), \sigma^2_\phi(\mathbb{W}^{(d)})\big),\\
    \bm{\nu}^{(d)} &\sim \pi_\psi(\bm{\nu}^{(d)}\mid \mathbf{z}^{(d)}) \equiv \prod_{k=1}^{K-1}\text{Kumaraswamy}\!\left(\nu^{(d)}_k; a_\psi(\mathbf{z}^{(d)}), b_\psi(\mathbf{z}^{(d)})\right).
\end{align*}

Since $\bm{\theta}^{(d)}$ is a deterministic function of $\bm{\nu}^{(d)}$, by the data-processing inequality~\citep{cover1999elements}, it follows that
\[
    \mathrm{KL}\!\left(q_{\phi}(\bm\theta^{(d)} \mid \mathbb{W}^{(d)}) \,\|\, p(\bm{\theta}^{(d)})\right) 
    \;\leq\; \mathrm{KL}\!\left(\pi_{(\phi, \psi)}(\mathbf{z}^{(d)},\bm{\nu}^{(d)} \mid \mathbb{W}^{(d)}) \,\|\, p(\mathbf{z}^{(d)},\bm{\nu}^{(d)})\right),
\]
where the joint variational distribution factorizes as
\(\pi_{(\phi,\psi)}(\mathbf{z}^{(d)},\bm{\nu}^{(d)}\mid\mathbb{W}^{(d)})=\pi_\phi(\mathbf{z}^{(d)}\mid\mathbb{W}^{(d)})\,\pi_\psi(\bm{\nu}^{(d)}\mid \mathbf{z}^{(d)})\),
and the prior as
\(p(\mathbf{z}^{(d)},\bm{\nu}^{(d)})=p(\mathbf{z}^{(d)})\,p(\bm{\nu}^{(d)})\).

Applying the chain rule of the KL divergence~\citep{cover1999elements}, we obtain
\begin{align*}
\text{KL}&\left(q_{\phi}(\bm\theta^{(d)} \mid \mathbb{W}^{(d)}) \mid\mid p(\bm{\theta}^{(d)})\right) \nonumber\\
& \leq \text{KL}\left(\pi_{(\phi, \psi)}(\mat{z}^{(d)},\bm{\nu}^{(d)} \mid \mathbb{W}^{(d)}) \mid\mid p(\mat{z}^{(d)},\bm{\nu}^{(d)})\right)\nonumber\\
& = \mathbb{E}_{\pi_{(\phi, \psi)}}\left [\log \frac{\pi_{(\phi, \psi)}(\mat{z}^{(d)},\bm{\nu}^{(d)}\mid\mathbb{W}^{(d)})}{p(\mat{z}^{(d)})p(\bm{\nu}^{(d)})}\right ]\nonumber\\
& = \mathbb{E}_{\pi_{(\phi, \psi)}}\left [
\log \frac{\pi_\phi(\mat{z}^{(d)}\mid\mathbb{W}^{(d)})\pi_\psi(\bm{\nu}^{(d)}\mid\mat{z}^{(d)})}{p(\mat{z}^{(d)})p(\bm{\nu}^{(d)})}
\right ]\nonumber\\
& = 
\mathbb{E}_{\pi_{(\phi, \psi)}}\left [
\log \frac{\pi_\phi(\mat{z}^{(d)}\mid\mathbb{W}^{(d)})}{p(\mat{z}^{(d)})} + \log\frac{\pi_\psi(\bm{\nu}^{(d)}\mid\mat{z}^{(d)})}{p(\bm{\nu}^{(d)})}
\right ]\nonumber\\
& = \mathbb{E}_{\pi_{\phi}}\left [
\log \frac{\pi_\phi(\mat{z}^{(d)}\mid\mathbb{W}^{(d)})}{p(\mat{z}^{(d)})}\right ] + \mathbb{E}_{\pi_{\phi}}
\left [ \mathbb{E}_{\pi_\psi}\left [\log\frac{\pi_\psi(\bm{\nu}^{(d)}\mid\mat{z}^{(d)})}{p(\bm{\nu}^{(d)})}
\right ]\right]\nonumber\\
& =
\text{KL}\left(\pi_\phi(\mat{z}^{(d)}\mid\mathbb{W}^{(d)})\mid\mid p(\mat{z}^{(d)})\right) + 
\mathbb{E}_{\pi_{\phi}}
\left [\text{KL}\left(\pi_\psi(\bm{\nu}^{(d)}\mid\mat{z}^{(d)})\mid\mid p(\bm{\nu}^{(d)})\right)\right]\nonumber\\
& =
\underbrace{\text{KL}\left(\pi_\phi(\mat{z}^{(d)}\mid\mathbb{W}^{(d)})\mid\mid \mathcal{N}(0, I)\right)}_{\text{KL}(\text{Gaussian}\mid\mid\text{Normal})} + 
\mathbb{E}_{\pi_{\phi}}
\left [\underbrace{\text{KL}\left(\pi_\psi(\bm{\nu}^{(d)}\mid\mat{z}^{(d)})\mid\mid \text{Beta}(a, b)\right)}_{\text{KL}(\text{Kumaraswamy}\mid\mid\text{Beta})}\right]\label{eq:loss3}.
\end{align*}

The KL between a Kumaraswamy and a Beta distribution has a closed form~\citep{nalisnick2016stick}. The outer expectation over $\pi_\phi$ is estimated by Monte Carlo sampling. 
Finally, plugging these components into the evidence lower bound, the used loss takes the form
\begin{align*}
\mathcal{L}(\bm\alpha, \bm\rho) &\geq\widehat{\mathcal{L}}(\bm\alpha, \bm\rho)  \nonumber\\
& \equiv \sum\limits_{d=1}^D \mathbb{E}_{q_{\phi}(\bm\theta^{(d)} \mid \mathbb{W}^{(d)})}
\left[\log p(\mathbb{W}^{(d)} \mid \bm{\theta}^{(d)}, \bm\alpha, \bm\rho)\right] - \text{KL}\left(q_{\phi}(\bm\theta^{(d)} \mid \mathbb{W}^{(d)}) \mid\mid p(\bm{\theta}^{(d)})\right)\nonumber\\
& \geq 
\sum\limits_{d=1}^D \mathbb{E}_{q_{\phi}(\bm\theta^{(d)} \mid \mathbb{W}^{(d)})}
\left[\log p(\mathbb{W}^{(d)} \mid \bm{\theta}^{(d)}, \bm\alpha, \bm\rho)\right]
-\text{KL}\left(\pi_\phi(\mat{z}^{(d)}\mid\mathbb{W}^{(d)})\mid\mid \mathcal{N}(0, I)\right) \nonumber\\
& \quad\quad - \mathbb{E}_{\pi_{\phi}}
\left [\text{KL}\left(\pi_\psi(\bm{\nu}^{(d)}\mid\mat{z}^{(d)})\mid\mid \text{Beta}(a, b)\right)\right].
\end{align*}

\begin{algorithm}[t]
\caption{Topic modeling at time $t$ with stick-breaking variational inference}
\label{alg:training-sbstream}
\begin{algorithmic}[1]
\State Initialize model and  variational parameters
\For{iteration $i=1,2,\dots$}
\State Compute $\beta_k = \softmax(\bm\rho^{\top}\alpha_k)$ for each topic $k$
\State Choose a minibatch $\mathcal{B}$ of documents
\ForAll{each document $\mathbb{W}^{(d)} \in \mathcal{B}$}
    \State Get normalized bag-of-word  $\mathbf{W}^{(d)}$
    \State Compute $\bm\mu^{(d)} = NN(\mathbf{W}^{(d)};\phi_\mu)$
    \State Compute $\bm\Sigma^{(d)} = NN(\mathbf{W}^{(d)};\phi_\Sigma)$
    \For{iteration $s=1,\dots, S$}
      \State $\epsilon\sim \mathcal N(0,I)$
      \State $\mat{z}^{(d)} \gets \bm\mu^{(d)} + \bm\sigma^{(d)} \odot \epsilon$
      \State Compute $a^{(d)} = NN(\mat{z}^{(d)}; \psi_a)$
      \State Compute $b^{(d)} = NN(\mat{z}^{(d)}; \psi_b)$
      \For{each topic $k\in\{1,\dots,K-1\}$}
        \State $u_k^{(d)} \sim \mathcal U(0,1)$
        \State $\nu_k^{(d)} \gets \big(1 - (1-u_k^{(d)})^{1/b_k^{(d)}}\big)^{1/a_k^{(d)}}$
      \EndFor
      \State Compute $\theta^{(d)}_1 \gets \nu^{(d)}_1$
      \For{iteration $k=2, \dots, K-1$}
        \State Compute $\theta^{(d)}_k \gets \nu^{(d)}_k \cdot \prod_{j=1}^{k-1} (1-\nu^{(d)}_j)$
      \EndFor
      \State Compute $\theta^{(d)}_K \gets \prod_{j=1}^{K-1} (1-\nu^{(d)}_j)$
      \State Compute $\widehat{\mat{W}}_s^{(d)} \gets \softmax({\bm\theta^{(d)}}\bm\beta^{\top})$
    \EndFor
    \State Compute $\widehat{\mat{W}}^{(d)}\gets\frac{1}{S}\sum_s\widehat{\mat{W}}_s^{(d)}$
    \State Estimate the ELBO in~\cref{eq:loss_final} and its gradient (backpropagation)
    \State Update model parameters $\bm\alpha$
    \State Update variational parameters ($\phi_\mu,\phi_\Sigma, \psi_a, \psi_b$)
  \EndFor
\EndFor
\end{algorithmic}
\end{algorithm}

\subsection{Learning Algorithm}
We present in~\Cref{alg:training-sbstream} the pseudocode of the learning algorithm at time step~$t$ for the SB-ETM model. In our experiments, we set $s = 1$.

\begin{figure}[htbp]
    \centering
    \includegraphics[width=\columnwidth]{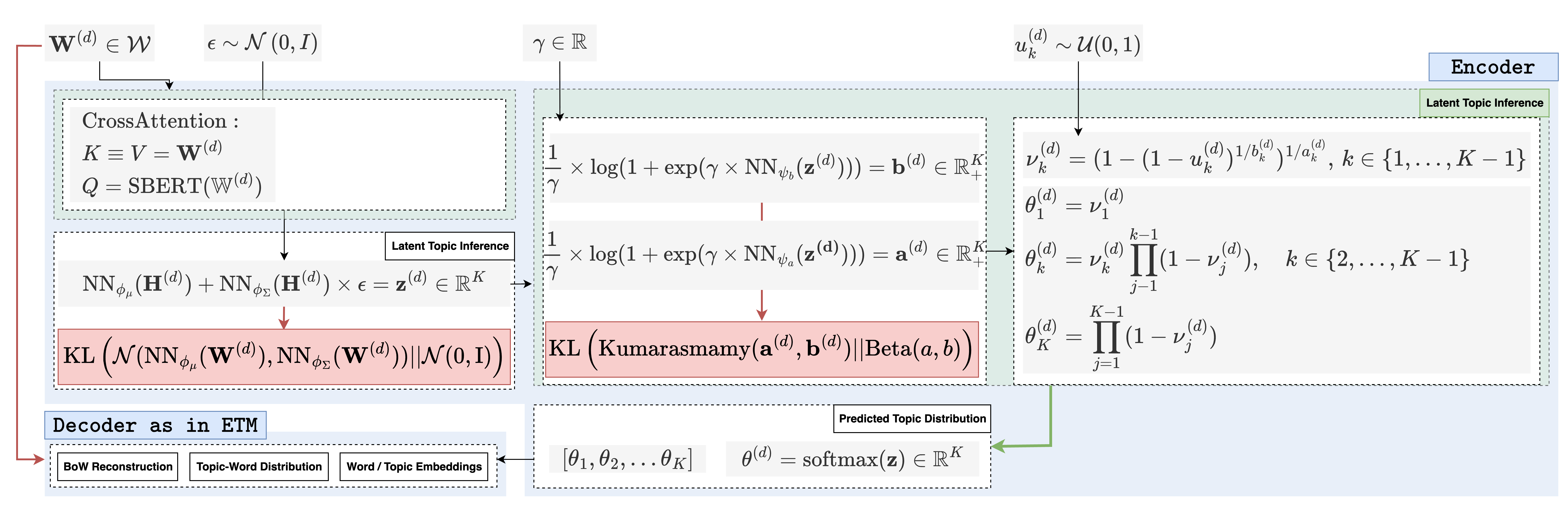}
    \caption{\textbf{Overall structure of SB-ETM at timestep $t$}. Green components denote the additional modules introduced beyond the original StreamETM/ETM design.}
    \label{fig:architecture}
\end{figure}
\begin{table}[h]
\caption{\textbf{Training time for each method considered in the paper}. The reported values correspond to the average training time per timestep; therefore, the total training time for a complete execution should be multiplied by 11 timesteps for \texttt{20kNewsGroupOnline}. The training time of {\sbstream} with the COT-based merging strategy and $K_\texttt{init}=50$ on \texttt{UkRuWarNews22-23} was 9.39 $\pm$ 1.65 minutes per timestep (to be multiplied by 22, the number of timesteps). We mark with a $*$ the cases where additional computational time was incurred due to server-related overhead.}
\centering
\begin{tabular}{r|c|c|c||c}
\toprule
Model & SBERT & OT & $K_{\texttt{init}}$ & Time (minutes)\\
\midrule
{\sbstream}  &  $\checkmark$ & Continuous & 15 & 15.11 $\pm$ 3.47\\
{\sbstream}  &  $\checkmark$ & Continuous & 25 & 15.04 $\pm$ 3.54\\
{\sbstream}  &  $\checkmark$ & Continuous & 50 & 18.42$^*$ $\pm$ 4.31\\
{\sbstream}  &  $\checkmark$ & Continuous & 75 & 15.47 $\pm$ 3.69\\
\midrule
{\sbstream}  &  $\checkmark$ & Discrete & 15 &
15.27 $\pm$ 3.48\\
{\sbstream}  &  $\checkmark$ & Discrete & 25 &
16.44 $\pm$ 3.57\\
{\sbstream}  &  $\checkmark$ & Discrete & 50 &
15.33 $\pm$ 3.65\\
{\sbstream}  &  $\checkmark$ & Discrete &  75 & 16.58 $\pm$ 3.65\\
\midrule
{\sbstream}  &  $\times$ & Continuous & 15 & 6.31 $\pm$ 0.79 \\
{\sbstream}  &  $\times$ & Continuous & 25 & 7.38 $\pm$ 1.11\\
{\sbstream}  &  $\times$ & Continuous & 50 & 7.19 $\pm$ 1.1\\
{\sbstream}  &  $\times$ & Continuous &  75 & 6.75 $\pm$ 1.0\\
\midrule
{\sbstream}  &  $\times$ & Discrete & 15 & 6.52 $\pm$ 0.99\\
{\sbstream}  &  $\times$ & Discrete & 25 & 6.25 $\pm$ 0.93\\
{\sbstream}  &  $\times$ & Discrete & 50 & 7.06 $\pm$ 1.03\\
{\sbstream}  &  $\times$ & Discrete & 75 & 6.58 $\pm$ 0.98\\
\midrule\midrule
StreamETM  &  $\times$ & Continuous & 3 & 0.85 $\pm$ 0.1\\
StreamETM  &  $\times$ & Continuous & 15 & 0.85 $\pm$ 0.1\\
StreamETM  &  $\times$ & Continuous & 25 & 0.86 $\pm$ 0.1\\
StreamETM  &  $\times$ & Continuous & 50 & 0.94 $\pm$ 0.11\\
StreamETM  &  $\times$ & Continuous &  75 & 0.91 $\pm$ 0.11\\
\midrule
StreamETM  &  $\times$ & Discrete &  3 & \textbf{0.88} $\pm$ 0.1\\
StreamETM  &  $\times$ & Discrete &  15 & 5.21$^*$ $\pm$ 20.22\\
StreamETM  &  $\times$ & Discrete & 25 &
0.91 $\pm$ 0.1\\
StreamETM  &  $\times$ & Discrete &  50 & 0.9 $\pm$ 0.12\\
StreamETM  &  $\times$ & Discrete &  75 & 0.93 $\pm$ 0.12\\
\bottomrule
\end{tabular}
\label{tab:time}
\end{table}
\begin{table*}[t]
    \centering
    \caption{\textbf{Ablation study.}}
    \begin{tabular}{r|c|c||c||c|c|c}
    \toprule
    Model & SBERT & OT & H$_{\text{(TC,TD)}}$ & $K_{\texttt{pred}}$ & $K_{\texttt{init}}$ & $K_{\texttt{real}}$ \\
\midrule
{\sbstream}  &  $\checkmark$ & Continuous & 0.88 $\pm$ 0.04 & 4.54 $\pm$ 0.68 & 15 & 3.18 $\pm$ 0.39 \\
{\sbstream}  &  $\checkmark$ & Continuous & 0.87 $\pm$ 0.04 & 4.53 $\pm$ 0.58 & 25 & 3.18 $\pm$ 0.39 \\
{\sbstream}  &  $\checkmark$ & Continuous & 0.88 $\pm$ 0.03 & 4.64 $\pm$ 0.56 & 50 & 3.18 $\pm$ 0.39 \\
{\sbstream}  &  $\checkmark$ & Continuous & 0.88 $\pm$ 0.04 & 4.51 $\pm$ 0.56 & 75 & 3.18 $\pm$ 0.39 \\
\midrule
{\sbstream}  &  $\checkmark$ & Discrete & 0.86 $\pm$ 0.06 & 5.57 $\pm$ 1.97 & 15 & 3.18 $\pm$ 0.39 \\
{\sbstream}  &  $\checkmark$ & Discrete & 0.87 $\pm$ 0.06 & 5.71 $\pm$ 2.29 & 25 & 3.18 $\pm$ 0.39 \\
{\sbstream}  &  $\checkmark$ & Discrete & 0.84 $\pm$ 0.07 & 5.42 $\pm$ 1.62 & 50 & 3.18 $\pm$ 0.39 \\
{\sbstream}  &  $\checkmark$ & Discrete & 0.84 $\pm$ 0.08 & 5.53 $\pm$ 1.97 & 75 & 3.18 $\pm$ 0.39 \\
\midrule
{\sbstream}  &  $\times$ & Continuous & 0.88 $\pm$ 0.03 & 4.55 $\pm$ 0.6 & 15 & 3.18 $\pm$ 0.39 \\
{\sbstream}  &  $\times$ & Continuous & 0.87 $\pm$ 0.03 & 4.24 $\pm$ 0.43 & 25 & 3.18 $\pm$ 0.39 \\
{\sbstream}  &  $\times$ & Continuous & 0.89 $\pm$ 0.03 & 3.89 $\pm$ 0.35 & 50 & 3.18 $\pm$ 0.39 \\
{\sbstream}  &  $\times$ & Continuous & 0.88 $\pm$ 0.03 & 4.29 $\pm$ 0.69 & 75 & 3.18 $\pm$ 0.39 \\
\midrule
{\sbstream}  &  $\times$ & Discrete & 0.86 $\pm$ 0.04 & 5.6 $\pm$ 1.31 & 15 & 3.18 $\pm$ 0.39 \\
{\sbstream}  &  $\times$ & Discrete & 0.88 $\pm$ 0.04 & 5.2 $\pm$ 1.79 & 25 & 3.18 $\pm$ 0.39 \\
{\sbstream}  &  $\times$ & Discrete & 0.84 $\pm$ 0.05 & 5.67 $\pm$ 1.71 & 50 & 3.18 $\pm$ 0.39 \\
{\sbstream}  &  $\times$ & Discrete & 0.83 $\pm$ 0.07 & 5.66 $\pm$ 1.67 & 75 & 3.18 $\pm$ 0.39 \\
\midrule\midrule
StreamETM  &  $\times$ & Continuous & 0.9 $\pm$ 0.04 & 2.02 $\pm$ 0.14 & 3 & 3.18 $\pm$ 0.39 \\
StreamETM  &  $\times$ & Continuous & 0.7 $\pm$ 0.13 & 8.39 $\pm$ 0.7 & 15 & 3.18 $\pm$ 0.39 \\
StreamETM  &  $\times$ & Continuous & 0.42 $\pm$ 0.14 & 14.22 $\pm$ 1.14 & 25 & 3.18 $\pm$ 0.39 \\
StreamETM  &  $\times$ & Continuous & 0.35 $\pm$ 0.28 & 28.0 $\pm$ 2.29 & 50 & 3.18 $\pm$ 0.39 \\
StreamETM  &  $\times$ & Continuous & 0.25 $\pm$ 0.26 & 42.8 $\pm$ 3.84 & 75 & 3.18 $\pm$ 0.39 \\
\midrule
StreamETM  &  $\times$ & Discrete & 0.93 $\pm$ 0.04 & 2.01 $\pm$ 0.1 & 3 & 3.18 $\pm$ 0.39 \\
StreamETM  &  $\times$ & Discrete & 0.65 $\pm$ 0.1 & 12.2 $\pm$ 2.22 & 15 & 3.18 $\pm$ 0.39 \\
StreamETM  &  $\times$ & Discrete & 0.3 $\pm$ 0.08 & 45.59 $\pm$ 10.3 & 25 & 3.18 $\pm$ 0.39 \\
StreamETM  &  $\times$ & Discrete & 0.05 $\pm$ 0.02 & 196.61 $\pm$ 83.19 & 50 & 3.18 $\pm$ 0.39 \\
StreamETM  &  $\times$ & Discrete & 0.03 $\pm$ 0.01 & 281.21 $\pm$ 85.52 & 75 & 3.18 $\pm$ 0.39 \\
\bottomrule
    
    \end{tabular}
    \label{tab:quantitative}
\end{table*}

\begin{table}[t]
\centering
\caption{\textbf{Thematic categories in \texttt{UkRuWarNews22-23}}.}
\begin{tabularx}{\linewidth}{l X | c}
\toprule
\textbf{Code} & \textbf{Description} & \textbf{N.\ Samples} \\
\midrule
\texttt{IDT}  & Identity, nationality and culture & 33 \\
\texttt{CDG}  & War crimes (including population displacements, indiscriminate bombings, treatment of bodies, civilian hostages, etc.) & 149 \\
\texttt{ENQU} & Investigations into war crimes & 55 \\
\texttt{GEN}  & Massacre / genocide & 60 \\
\texttt{REEC} & Rewriting of history (including the issue of Russian territory and the question of Nazism) & 52 \\
\texttt{TERR} & Terrorism & 28 \\
\texttt{PRO}  & Propaganda & 49 \\
\texttt{NUC}  & Nuclear security / management of energy resources & 60 \\
\texttt{OCC}  & Western aid & 50 \\
\texttt{MIGR} & Migration crisis and ``quality'' of migrants & 60 \\
\texttt{OTAN} & NATO enlargement & 57 \\
\texttt{POL}  & Positioning of politicians & 39 \\
\texttt{PPU}  & Positioning of public figures & 18 \\
\texttt{ONG}  & Positioning and actions of NGOs & 27 \\
\texttt{PAIX} & Question of the conditions of peace & 65 \\
\texttt{INFO} & Information warfare & 41 \\
\texttt{REL}  & International relations & 26 \\
\texttt{HUMA} & Humanitarian aid & 34 \\
\texttt{FUIT} & Brain drain and artists' flight & 20 \\
\texttt{ART}  & Artists in the war & 35 \\
\texttt{BOYC} & Question of the boycott of Russian art & 8 \\
\texttt{BOY}  & Boycotting & 13 \\
\texttt{CIO}  & Debates on the 2024 Olympics & 34 \\
\texttt{CYBR} & Cyberconflict & 29 \\
\texttt{HACK} & Hacktivism & 35 \\
\texttt{MERC} & Mercenaries & 30 \\
\texttt{VOL}  & Volunteers & 37 \\
\texttt{AIR}  & Aerial warfare & 36 \\
\texttt{MISS} & Missiles & 38 \\
\texttt{DRO}  & Drones & 35 \\
\texttt{DISC} & Political speeches & 993 \\
\texttt{PURG} & Purge of generals & 21 \\
\texttt{CRIM} & Return of criminals amnestied by their engagement in Ukraine & 24 \\
\texttt{OLIG} & Disappearances of Russian oligarchs & 32 \\
\texttt{JOUR} & Atrocities against journalists on the front & 25 \\
\texttt{EDUC} & Militarization of children & 22 \\
\texttt{SANT} & Health of Vladimir Putin & 10 \\
\texttt{RESI} & Individual acts of resistance by Russian soldiers (refusal to fight and desertion) & 27 \\
\texttt{PLAI} & Complaints by Russian soldiers and their families & 23 \\
\bottomrule
\end{tabularx}
\label{tab:topic-category-codes}
\end{table}

\section{ADDITIONAL MATERIAL TO SECTION~\ref{sec:exp_setting}}
\label{app:preprocessing}
\subsection{Datasets}
\paragraph{\texttt{20kNewsGroupOnline}.} \Cref{fig:ground_truth} provides the overall picture of the topic frequencies over time as constructed in~\citep{granese2025merging}. Following their strategy, we randomly draw 8 times approximately 5k samples from the total datasets. 
We partition the 5k samples into 500 sample batches to simulate an 11-step scenario.

\paragraph{\texttt{UkRuWarNews22-23}.} 
The corpus comprises 4.211 news articles on the Russia–Ukraine war from 2022 to 2023. We focus on the 2.430 articles written in English, which span 39 thematic categories\footnote{In the main body of the paper, we misindicate the sample count as 2.652 and 1.617; the correct values are 4.211 and 2.430, respectively. This affects only the text in~\Cref{sec:exp_setting}; all computations, tables, and figures use 4.211 and 2.430.}
The dataset comprises 840 articles in 2022 and 1590 in 2023.
For each dataset entry, the article URL, publication (or last-modified) date, and cleaned article text are included. All articles were manually reviewed to correct errors and fill missing fields, and the text was thoroughly cleaned to remove non-content elements (e.g., advertisements, newsletter prompts, image captions, alternative text, and cookie banners).
The average number of tokens per article is $\sim$426 (max 4038, min 7). The average number of characters per article is $\sim$3.125 (max 31.657, min 46).
The English corpus covers 113 news outlets, with president.gov.ua, Reuters, AP News, BBC, The Guardian, and CNN Edition among the most represented sources. The articles have been categorised by three experts in political and communication sciences.
The reference to the dataset will be provided upon acceptance to ensure anonymity.

\paragraph{Text preprocessing.}
We process the articles with a uniform pipeline. 
\begin{itemize}
    \item\underline{Lemmatization:} documents are lemmatized with \texttt{spaCy} large models (\texttt{en\_core\_web\_lg}), disabling parser/NER to retain only lemmatization. 
    \item\underline{Normalization and tokenization:} we lowercase, replace apostrophes with spaces, strip punctuation, and tokenize with NLTK’s \texttt{word\_tokenize}. 
    \item\underline{Token filtering:} we keep alphabetic tokens of length $>2$ and remove language-specific stopwords (NLTK lists), augmenting the English list with a small domain set \{\texttt{notoc}, \texttt{coxnet}\}. 
    \item\underline{Vocabulary pruning:} we remove tokens with frequency $\le 1$ and drop extremely frequent tokens using a document-frequency threshold (appear in $>70\%$ of documents). 
\end{itemize}
The preprocessing pipeline is applied \emph{independently at each timestep} $t$: pruning of low- and high-frequency tokens is computed within the current batch only, not over the full corpus. The same procedure is used for both \texttt{20kNewsGroupOnline} and \texttt{UkRuWarNews22-23}. For \texttt{20kNewsGroupOnline}, we additionally strip e-mail headers (contact fields) and subject lines before processing.

\subsection{Model Architecture and Training Procedure}
We extend the StreamETM architecture, which builds upon the Embedded Topic Model, by integrating two additional modules. 
The base architecture consists of two embedding matrices: a \texttt{word embedding matrix}, initialized from pre-trained \texttt{GloVe} vectors of dimension 300 and kept fixed during training, and a \texttt{topic embedding matrix} (\texttt{Xavier} initialization), of the same dimensionality, which remains fully trainable. These matrices form a shared latent space connecting topics and words.

The encoder block of the main body architecture consists of a feedforward bag-of-words network that maps each normalized document vector into a dense hidden representation, combined with a Cross-Attention Encoder (cf. Sentence-BERT module) that integrates sentence-level representations extracted from a multilingual Sentence-BERT model (\texttt{paraphrase-multilingual-mpnet-base-v2}). The fused embedding is normalized through LayerNorm and combined with the BoW representation via a residual connection. Two fully connected latent heads ($\text{NN}_{\phi_{\mu}}(\cdot)$, $\text{NN}_{\phi_{\Sigma}}(\cdot)$) produce the mean and log-variance of the Gaussian posterior, each followed by Batch Normalization for stability. The reparameterized latent sample is then processed by a Stick-Breaking Encoder (cf. Stick-Breaking module), which maps it into document–topic proportions. The decoder reconstructs the document’s word distribution by projecting topic proportions onto the shared embedding space, followed by Batch Normalization and a softmax layer.

Optimization uses Adam with an initial learning rate of 0.01 (One-Cycle schedule, linearly increasing during the first 10\% of the total steps and then decaying following a cosine annealing strategy), weight decay of 0.006, batch size of 1024, and a scheduler with a step size of 5000, over 2600 epochs. All experiments were run on NVIDIA H100 GPUs.

\paragraph{Sentence-BERT (SBERT) module.} The Cross-Attention Encoder fuses lexical and contextual information by aligning bag-of-words and sentence-level embeddings in a shared latent space. It consists of three components:
\begin{itemize}
    \item\underline{Projection layers}: two linear mappings project the BoW vector and the SBERT embeddings into the same latent dimensionality.
    \item\underline{Multi-head attention}: a single-head MultiheadAttention module (batch-first) takes the SBERT embeddings as queries and the projected BoW representation as both keys and values. 
    \item\underline{Aggregation}: the attention matrix is passed through a LayerNorm operation, producing the final document embedding. This embedding is then added to the original BoW embedding computed by the feedforward bag-of-words network.
\end{itemize}
\paragraph{Stick-Breaking module.} The Stick-Breaking Encoder converts the latent Gaussian variables into normalized document–topic proportions through a differentiable stick-breaking process parameterized by the Kumaraswamy distribution. The module comprises two parallel linear layers, each mapping the latent vector to per-topic shape parameters ($\text{NN}_{\psi_{a}}(\cdot)$, $\text{NN}_{\psi_{b}}(\cdot)$). The outputs are activated through softplus and stabilized by a small additive constant. Stick variables are sampled via a deterministic Kumaraswamy sampler using clamped uniform noise to ensure numerical robustness. Topic weights are then obtained through cumulative products over the remaining stick lengths, resulting in a simplex-valued vector. 
We apply a warm-starting from the model at the previous timestep. Overlapping weights are copied, while new parameters are initialized through Xavier initialization; bias terms for newly added topics are adjusted to emphasize their activation. 

We provide in~\Cref{fig:architecture} an overview of the SB-ETM model at time $t$.

\begin{figure}[t]
    \centering
    \begin{subfigure}{\columnwidth}
    \includegraphics[width=\columnwidth]{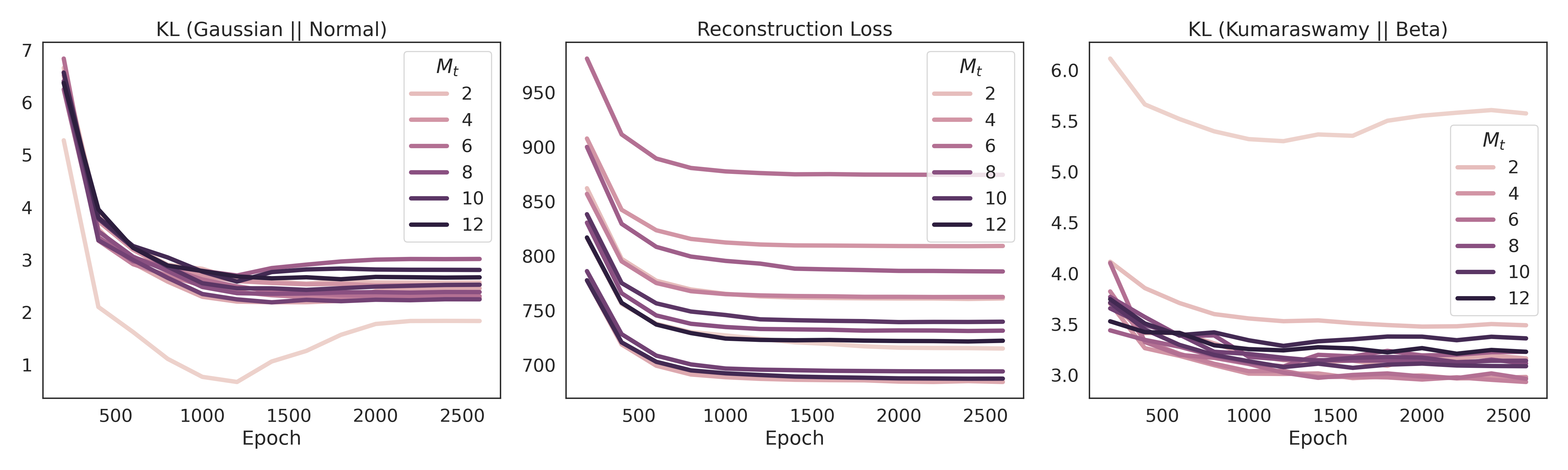}
    \caption{\texttt{20kNewsGroupOnline}.}
    \end{subfigure}
    \begin{subfigure}{\columnwidth}
    \includegraphics[width=\columnwidth]{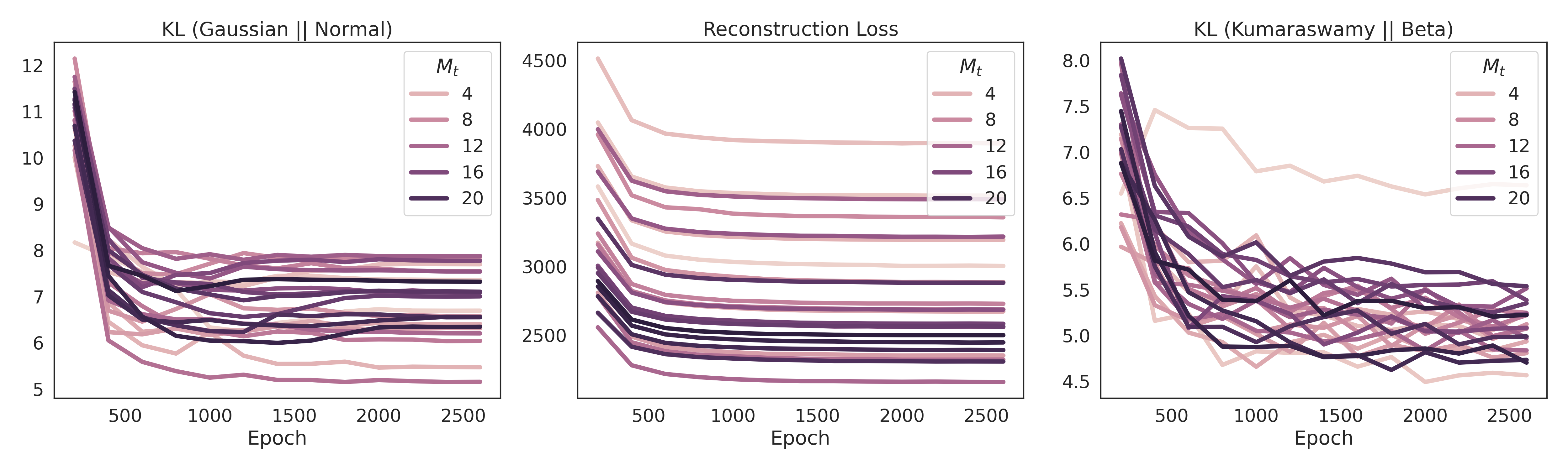}
    \caption{\texttt{UkRuWarNews22-23}.}
    \end{subfigure}
    \caption{\textbf{Losses evolution during training}. {\sbstream} (COT-based merging strategy and $K_{\texttt{init}}=50$).}
    \label{fig:losses}
\end{figure}

\paragraph{$\omega_R, \omega_G, \omega_S$ and the Beta distribution parameter during the training.}
We observe from~\Cref{tab:parameters} that the best-performing configurations correspond to cases where the value of $\omega_S$ is small compared to $\omega_G$, or more generally, when $\omega_S < 0.1$. Regarding the Beta parameterization, the most coherent results are obtained when $a = b = 0.5$, i.e., when the Beta distribution takes a U-shaped form. This behaviour aligns well with our stick-breaking formulation: at the beginning of the process, a higher probability mass is assigned to the first topics, while the model is progressively encouraged to allocate more weight to the newly introduced ones. Based on these findings, for all the experiments presented in this paper, we adopt the configuration highlighted in bold in the table. Moreover, we provide in~\Cref{fig:losses} the evolution of the 3 loss components during the training for each timestep on \texttt{20kNewsGroupOnline}. Reconstruction loss decreases and stabilizes near 700–750, while both KL terms converge to moderate values ($\approx$2–3).
This range reflects balanced regularization: the variational posteriors remain close to their priors, preventing collapse and preserving informative latent topics. Note that the reconstruction loss values are relatively high because they represent the sum of token-level log-likelihoods per document; they therefore scale with document length and corpus size, not with model inefficiency. This difference in scale is intrinsic to the ELBO formulation. A similar behaviour is also observed in \texttt{UkRuWarNews22-23}.

\paragraph{Training time analysis.} \Cref{tab:time} reports the training time for each method considered in the paper. We observe that the SBERT module is the most computationally expensive component, as its inclusion approximately doubles the training time per timestep. Notably, neither $K_\texttt{init}$ nor the COT-based merging strategy introduces a significant computational overhead. In contrast, the Stick-Breaking module increases the training time by approximately six minutes compared to the original model (StreamETM - Discrete - $K_\texttt{init}=3$).
\begin{table}[H]
    \centering
    \caption{Quantitative results on \texttt{20kNewsGroupOnline} when  {\sbstream} (COT-based merging strategy and $K_\texttt{init}=50$) has been trained using different parameter values.}
    \begin{tabular}{c|c|c |c|c||c|c|c|c}
    \toprule
    $\omega_R$ & $\omega_G$ & $\omega_S$ & $a$ & $b$ & $\text{H}_{\text{(TC, TD)}}$ & $K_\texttt{pred}$ & $K_\texttt{init}$ & $K_\texttt{real}$\\
\midrule
    \textbf{1} & \textbf{1} & \textbf{0.05} & \textbf{0.5} & \textbf{0.5} & 0.97 $\pm$ 0.04 & 4.64 $\pm$ 0.57 & 50 & 3.18 $\pm$ 0.39 \\
    1 & 1 & 0.05 & 1 & 2 & 0.98 $\pm$ 0.05 & 4.82 $\pm$ 0.75 & 50 & 3.18 $\pm$ 0.39\\
    1 & 1 & 0.05 & 1 & 1& 0.94 $\pm$ 0.08 & 5.0 $\pm$ 0.77 & 50 & 3.18 $\pm$ 0.39 \\
    1 & 1 & 0.05 & 1 & 0.5 & 0.94 $\pm$ 0.08 & 5.0 $\pm$ 0.77 & 50 & 3.18 $\pm$ 0.39 \\
    1 & 1 & 0.05 & 2 & 2 & 0.95 $\pm$ 0.06 & 4.0 $\pm$ 0.0 & 50 & 3.18 $\pm$ 0.39\\
    1 & 0.5 & 0.1 & 0.5 & 0.5 & 0.92 $\pm$ 0.05 & 5.36 $\pm$ 1.21 & 50 & 3.18 $\pm$ 0.39\\
    1 & 1.5 & 0.01 & 0.5 & 0.5 & 0.97 $\pm$ 0.03 & 5.0 $\pm$ 0.0 & 50 & 3.18 $\pm$ 0.39\\
    1 & 0.1 & 0.2 & 0.5 & 0.5 & 0.95 $\pm$ 0.05 & 4.91 $\pm$ 0.3 & 50 & 3.18 $\pm$ 0.39 \\
    1 & 0.5 & 0.005 & 0.5 & 0.5 & 0.98 $\pm$ 0.03 & 5.91 $\pm$ 0.3 & 50 & 3.18 $\pm$ 0.39\\
\bottomrule
\end{tabular}
\label{tab:parameters}
\end{table}
\section{ADDITIONAL MATERIAL TO SECTION~\ref{sec:discussion}}
\label{app:discussion}
\subsection{Tracing the Topic Evolution}
\begin{algorithm}[t]
\caption{Association between of topic embeddings $\bm{\alpha}^{(t-1)}$ and $\bm{\alpha}^{(t)}$ at consecutive timesteps via transport-based matching at threshold $\varepsilon$}
\label{alg:tracing}
\begin{algorithmic}[1]
\State Compute $X \gets \begin{bmatrix} \bm{\alpha}^{(t-1)} \\ \bm{\alpha}^{(t)} \end{bmatrix}$    
\State Compute $\Sigma \gets \text{Cov}(X) + \epsilon_{\text{ridge}} \cdot \mathbf{I}_d$ \Comment{where $d = K_{\texttt{pred}}^{(t-1)} + K_{\texttt{pred}}^{(t)}$ and $\epsilon_{\text{ridge}}=1e-6$}
\State Let $\lambda_1 \leq \dots \leq \lambda_d$ be the eigenvalues of $\Sigma$; set $\lambda \gets \lambda_d$
\State Compute threshold $T \gets \sqrt{\lambda} \cdot \varepsilon$
\State Compute transport matrix $\text{OT} \gets \texttt{UnbalancedOT}(\bm{\alpha}^{(t)}, \bm{\alpha}^{(t-1)})$ \Comment{OT $\in \mathbb{R}^{K^{(t)} \times K^{(t-1)}}$}
\For{each topic $i \in \{1, \dots, K_{\texttt{pred}}^{(t)}\}$}
  \State $w_i \gets \max_j \text{OT}[i, j]$
  \State $\text{index}_i \gets \arg\max_j \text{OT}[i, j]$
  \If{$w_i \geq T$}
    \State Assign topic $i$ to class $\text{index}_i$
  \Else
    \State Flag topic $i$ as a new topic
  \EndIf
\EndFor
\end{algorithmic}
\end{algorithm}

The COT merging approach does not allow us to directly track the evolution of topics over time, as topics are directly projected into embedding spaces. We provide in~\Cref{alg:tracing} the pseudo code of the algorithm we used to trace the topic evolution over time.
To compute the transport map in the algorithm, we use the Cosine Distance for the cost map. The transport map is computed using the Python function \texttt{ot.unbalanced.mm\_unbalanced}, with KL divergence and marginal relaxation at 0.09. In the experiments, we set $\varepsilon=0.01$.

\subsection{Ablation Study} In~\Cref{tab:quantitative}, we report the complete set of quantitative results, evaluating the performance of the proposed model across all possible configurations—namely, with and without the SBERT module, and using both COT- and DOT-based merging strategies. The main trend that emerges concerns the DOT-based merging strategy, which generally leads to lower H$_\text{(TC, TD)}$ values compared to the COT-based strategy. As discussed earlier, this result is not surprising, since the COT strategy explicitly projects the topic embeddings at time $t$ into the latent space at time $t-1$.
When comparing versions with and without the SBERT module, we do not observe a clear quantitative advantage from its inclusion. This outcome is expected, as TC and TD are not directly sensitive to topic semantics: even if topics become semantically more coherent, these metrics only measure whether words co-occur within a given context window in the text. For this reason, we complement this analysis with a qualitative evaluation in the next section.

\subsection{Qualitative Results}
\begin{figure*}[t]
    \centering
    \begin{subfigure}[t]{0.8\columnwidth}
        \centering
        \includegraphics[width=\columnwidth]{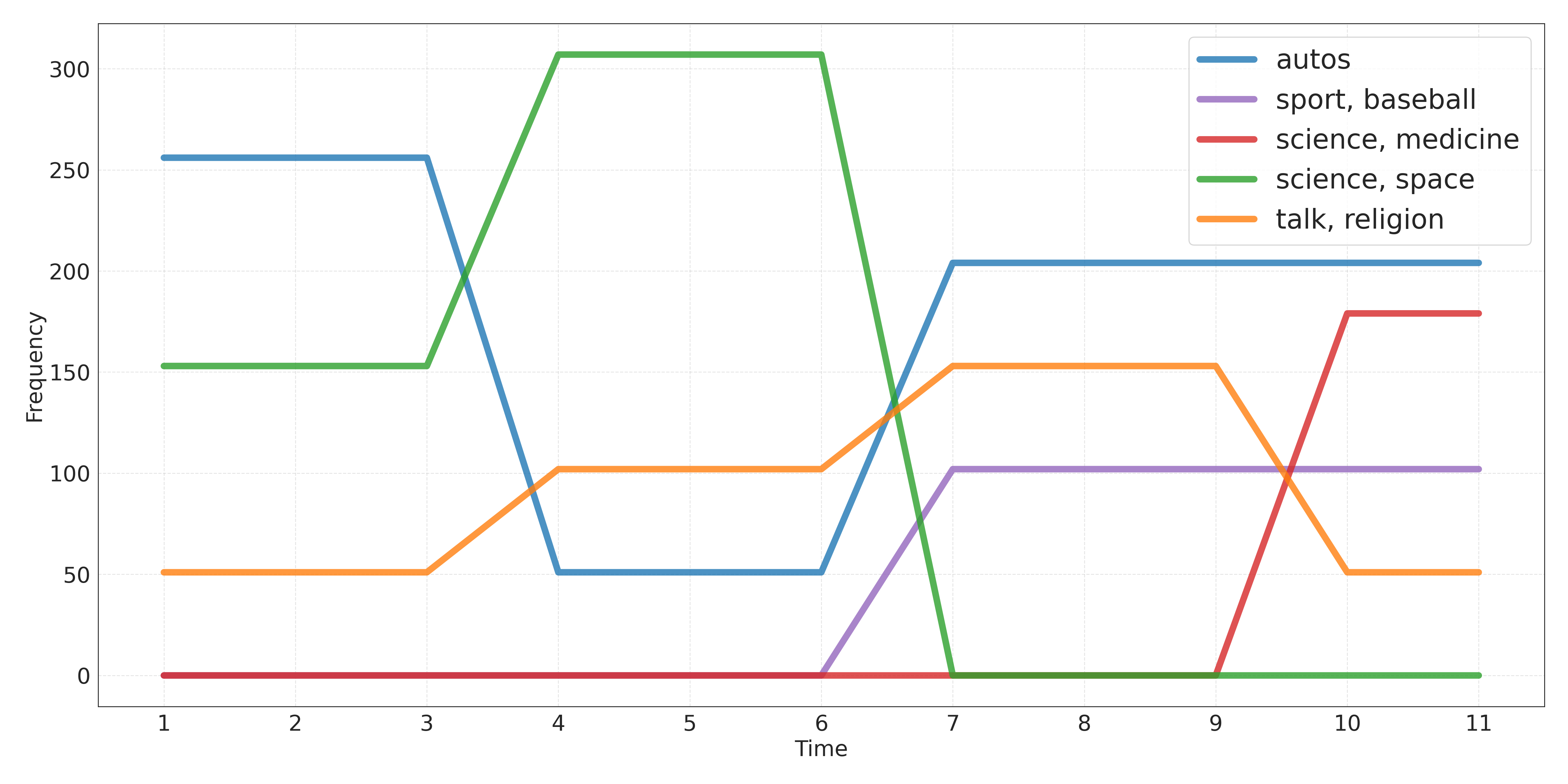}
        \caption{Ground Truth}
        \label{fig:ground_truth}
    \end{subfigure}%
    \hfill
    \begin{subfigure}[t]{0.8\columnwidth}
        \centering
        \includegraphics[width=\columnwidth]{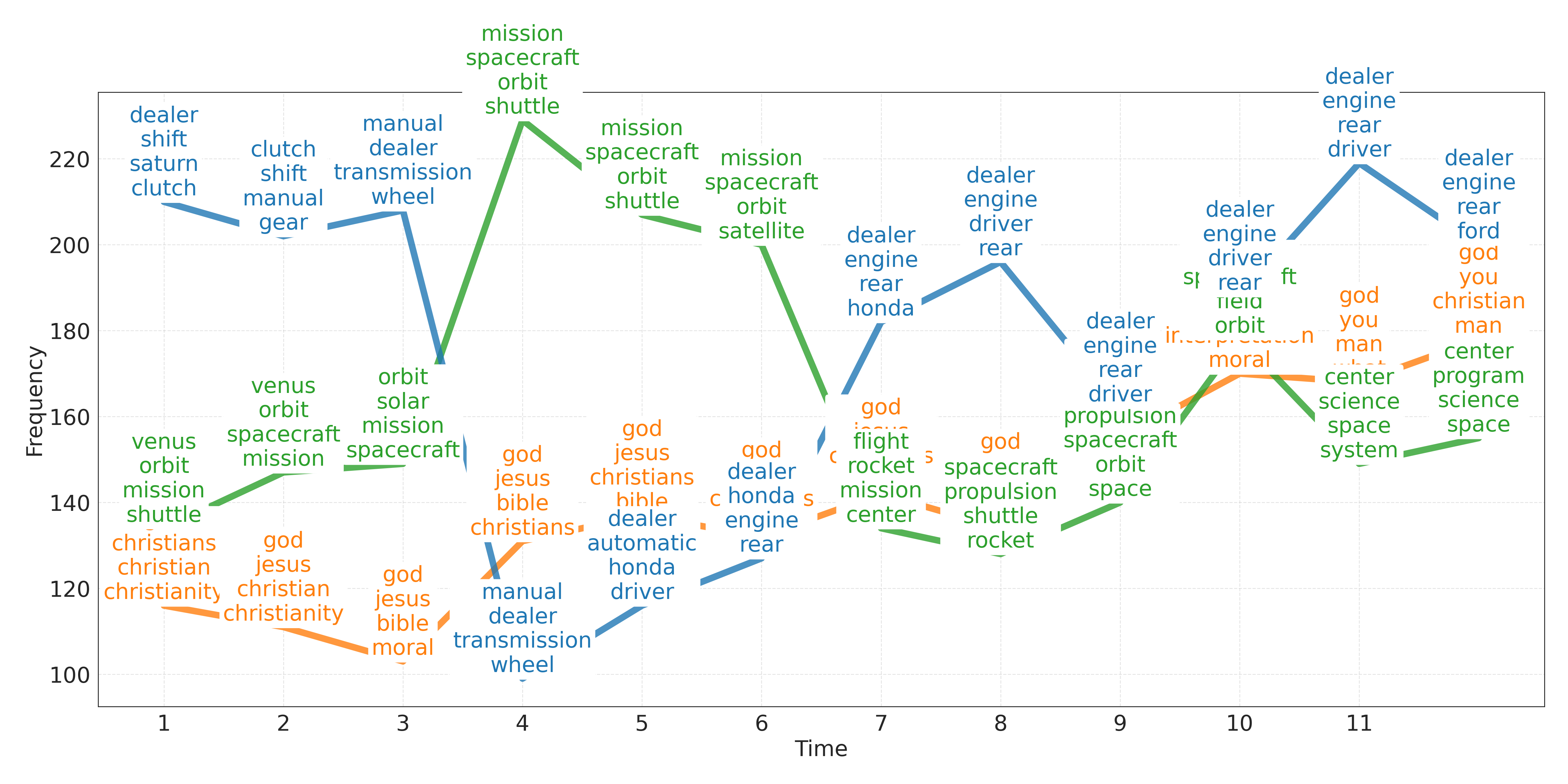}
        \caption{StreamETM}
        \label{fig:stream}
    \end{subfigure}
    \hfill
    \begin{subfigure}[t]{0.8\columnwidth}
        \centering
        \includegraphics[width=\columnwidth]{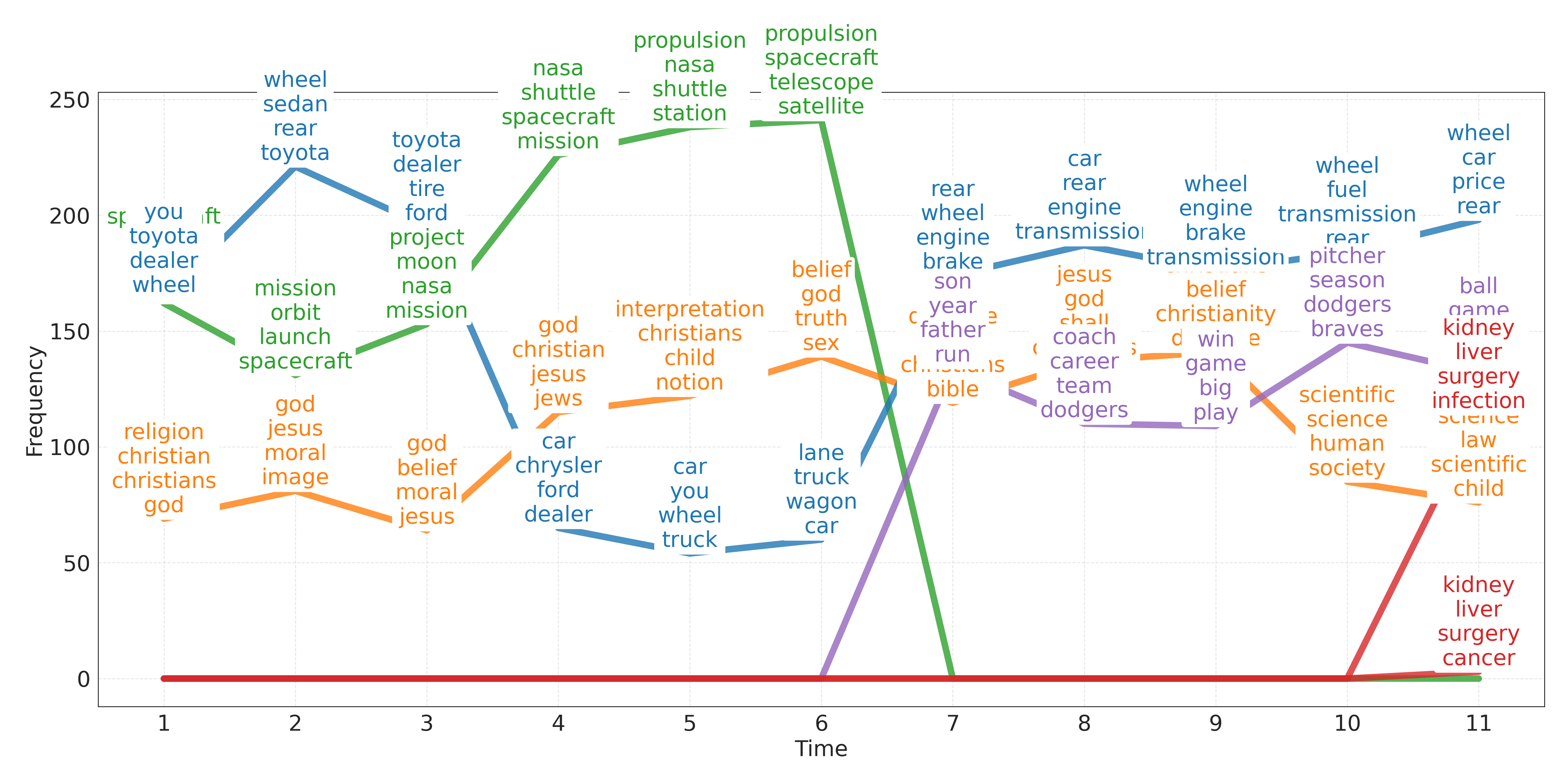}
        \caption{\sbstream}
        \label{fig:sbstream}
    \end{subfigure}
    \caption{\textbf{Topic frequencies over time, \texttt{20kNewsGroupOnline}}. In (b) StreamETM as in \citet{granese2025merging}: DOT-based merging, $K_{\texttt{init}}{=}3$. 
In (c) {\sbstream}: SBERT-enabled, COT-based merging, $K_{\texttt{init}}{=}50$.
}
    \label{fig:distribution_custom}
\end{figure*}

\paragraph{Topic frequency distributions on \texttt{20kNewsGroupOnline}.} \Cref{fig:distribution_custom} shows the topic frequency distributions for {\sbstream} (SBERT-enabled, COT-based merging, and $K_\texttt{init}=50$) and for StreamETM in its best-performing configuration (i.e., as in the seminal paper by~\cite{granese2025merging}, with DOT-based merging and $K_\texttt{init}=3$). Ideally, a well-functioning model should be able to reproduce the distribution in~\Cref{fig:ground_truth}. We observe that while both methods correctly capture the topics related to \texttt{science, space}, \texttt{autos}, and \texttt{talk, religion} up to timestep 7, StreamETM begins to confuse topics afterward, as the \texttt{science, space} category, in the ground truth reference, disappears, leaving room for the \texttt{sport, baseball} topic. This behavior, however, is correctly handled by our model, which also succeeds at timestep 10 in identifying the topic related to \texttt{science, medicine}, a topic that is instead absorbed into others by StreamETM.

\begin{figure}[t!]
    \centering
    
    \begin{subfigure}[t]{0.5\columnwidth}
        \centering
        \includegraphics[width=\columnwidth]{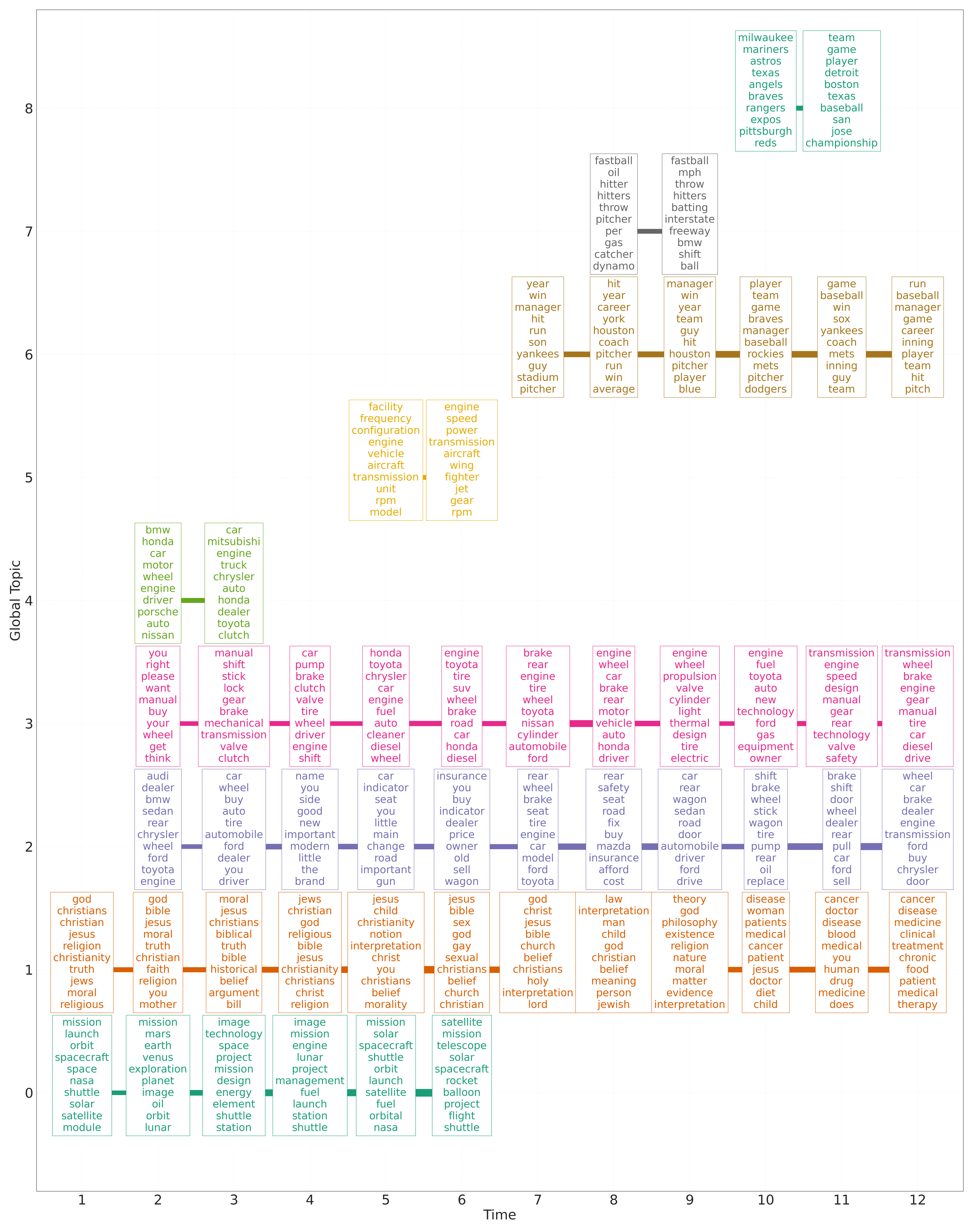}
        \caption{{\sbstream} without SBERT module (COT-merging)}
        \label{fig:interpretative_no_sentence_gauss}
    \end{subfigure}%
    \hfill
    \begin{subfigure}[t]{0.5\columnwidth}
        \centering
        \includegraphics[width=\columnwidth]{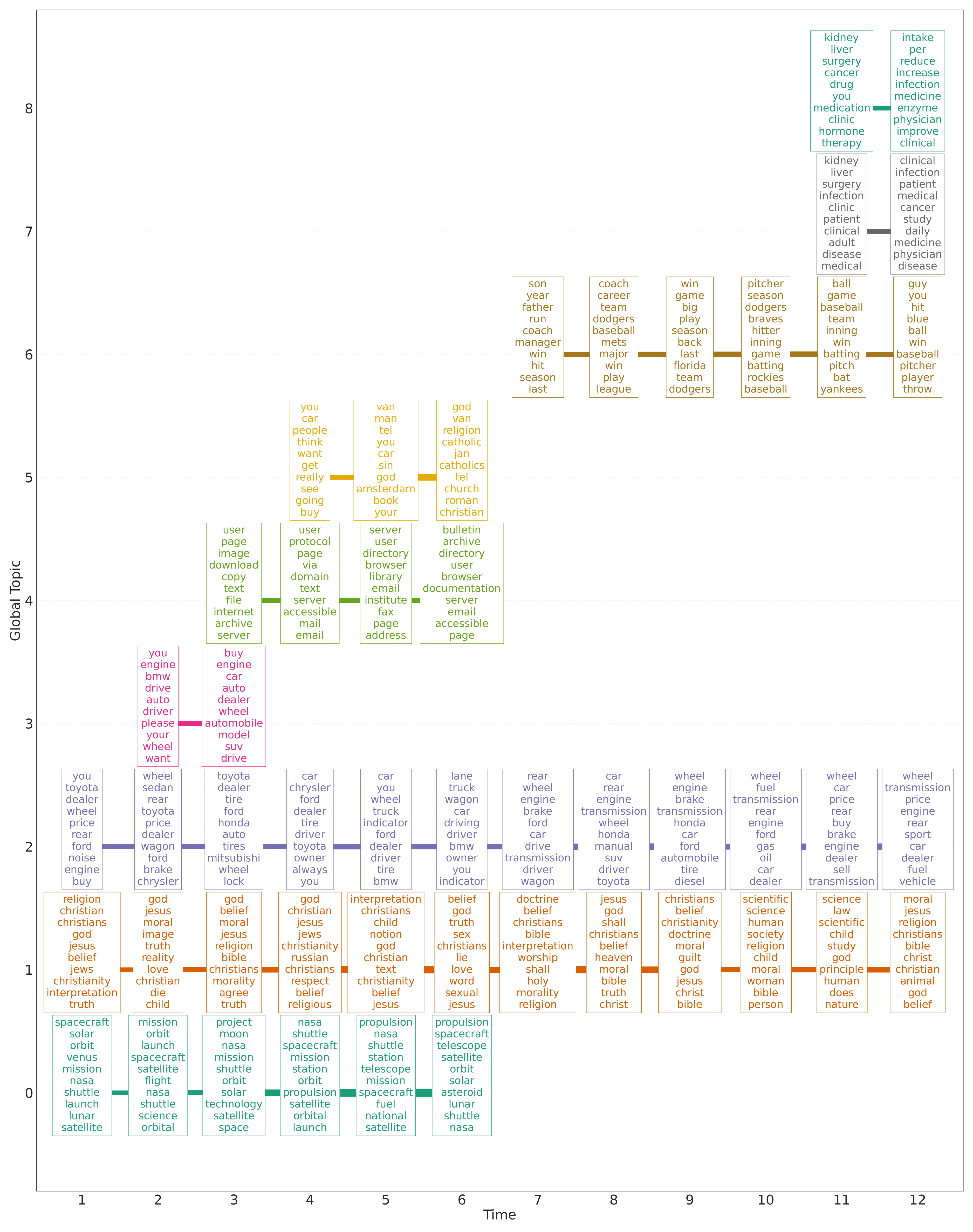}
        \caption{{\sbstream} with SBERT module (COT-merging)}
        \label{fig:interpretative_full_gauss}
    \end{subfigure}%
    \caption{\textbf{Qualitative results on \texttt{20kNewsGroupOnline}}. Both {\sbstream} models were trained under identical conditions with $K_\texttt{init}=50$, sharing the same seed and document sets. The line thickness between top-word boxes indicates the similarity between topics according to the tracing strategy.}
    \label{fig:interpretative}
\end{figure}
\paragraph{SBERT module as a semantic regularizer.} \Cref{fig:interpretative} compares the models trained with and without the SBERT module, both using the COT-based merging strategy. For each timestep, we plot the top words describing each topic and apply~\Cref{alg:tracing} to link the most similar topics over time.

From the plots, we observe that \textit{(i)} at timestep 10, in the model without SBERT, the topic related to religion (the orange one, index 1) gradually drifts in meaning and eventually becomes associated with medicine-related terms. In contrast, in the SBERT-enabled model, the religious topic remains stable across time, while a distinct topic emerges for medicine (indexes 7 and 8). \textit{(ii)} The model without SBERT also produces two parallel topics about automobiles—described by different sets of words but semantically overlapping—whereas this redundancy does not occur in the SBERT-based model.

Overall, these differences highlight the role of the SBERT module in enhancing the semantic consistency and temporal coherence of topics. By grounding the topic representations in a semantic embedding space, SBERT reduces noisy drift and prevents concept splitting or merging that occurs solely due to lexical variation. The resulting trajectories are smoother and more interpretable, with topics preserving their meaning over time while allowing the emergence of genuinely new semantic areas (e.g., the separation of religion and medicine).

\end{document}